\documentclass[manuscript, authorversion, nonacm]{acmart}

\AtBeginDocument{%
  }
\setcopyright{acmlicensed}
\copyrightyear{2026}
\acmYear{2026}
\acmDOI{XXXXXXX.XXXXXXX}
\acmJournal{HEALTH}
\acmVolume{TBD}
\acmNumber{TBD}
\acmArticle{TBD}

\usepackage{algorithm}
\usepackage{algorithmic}
\usepackage{subcaption}
\usepackage{amsmath}
\usepackage{booktabs}
\usepackage{multirow}
\usepackage{ulem}
\usepackage{comment}
\usepackage{graphicx}
\usepackage{amsfonts}
\usepackage{textcomp}
\usepackage{caption}
\usepackage{xcolor}

\newcommand{\figref}[1]{Fig.~\ref{fig:#1}}
\newcommand{\secref}[1]{Section~\ref{sec:#1}}
\newcommand{\tblref}[1]{Table~\ref{tbl:#1}}

\newcommand{\danets}{DAN\scalebox{0.8}{ETS} }

\begin{document}
\title{Use of What-if Scenarios to Help Explain Artificial Intelligence Models for Neonatal Health}

\author{Abdullah Mamun}
\authornote{Corresponding author.}
\orcid{0000-0001-8330-1383}
\affiliation{%
\department{College of Health Solutions and School of Computing and Augmented Intelligence}
  \institution{Arizona State University}
  \city{Phoenix}
  \state{AZ}
  \country{USA}
}
\email{a.mamun@asu.edu}

\author{Lawrence D. Devoe}
\orcid{0000-0003-1026-8388}
\affiliation{%
\institution{Medical College of Georgia at Augusta University}
\city{Augusta}
\state{GA}
\country{USA}}
\email{ldevoe@augusta.edu}

\author{Mark I. Evans}
\affiliation{%
\institution{Fetal Medicine Foundation of America}
\city{New York}
\state{NY}
\country{USA}}
\email{evans@compregen.com}

\author{David W. Britt}
\affiliation{%
\institution{Fetal Medicine Foundation of America}
\city{New York}
\state{NY}
\country{USA}}
\email{dwbrit01@me.com}

\author{Judith Klein-Seetharaman}
\orcid{0000-0002-4892-6828}
\affiliation{%
\department{College of Health Solutions}
  \institution{Arizona State University}
  \city{Phoenix}
  \state{AZ}
  \country{USA}
}
\email{judith.klein-seetharaman@asu.edu}

\author{Hassan Ghasemzadeh}
\orcid{0000-0002-1844-1416}
\affiliation{%
\department{College of Health Solutions}
  \institution{Arizona State University}
  \city{Phoenix}
  \state{AZ}
  \country{USA}
}
\email{hassan.ghasemzadeh@asu.edu}

\renewcommand{\shortauthors}{Mamun et al.}

\begin{abstract}
Early detection of intrapartum risks enables timely interventions to prevent or mitigate adverse labor outcomes such as cerebral palsy. However, accurate automated systems to support clinical decision-making during delivery are currently lacking. To address this gap, we propose Artificial Intelligence for Modeling and Explaining Neonatal Health (AIMEN), a deep learning framework that predicts adverse labor outcomes from maternal, fetal, obstetrical, and intrapartum factors while providing interpretable reasoning behind its predictions. AIMEN reveals how specific modifications to input variables could alter predicted outcomes, enhancing clinical insight. To address class imbalance and limited sample size, AIMEN employs Conditional Tabular GAN (CTGAN) for data augmentation. This process includes synthetic data generation, and we investigate in detail properties such as relaxing feature bounds for a subset of training points to explore slightly out-of-range physiological values, and applying silhouette-score-based filtering to increase the separability of synthetic samples. AIMEN uses an ensemble of fully connected neural networks for classification and outperforms state-of-the-art models such as XGBoost, TabNet, DANet, and LightGBM, achieving an average F1 score of 0.784 in predicting high-risk deliveries. Moreover, AIMEN generates counterfactual explanations that identify actionable changes involving only two to three attributes on average. Resources: https://github.com/ab9mamun/AIMEN.

\end{abstract}



\maketitle

\textbf{\small Accepted for publication in ACM Transactions on Computing for Healthcare (ACM HEALTH), April 2026.}

\section{Introduction}
\label{sec:intro}
\begin{figure}[tbh!]
    \centering
        \includegraphics[width=0.48\textwidth]{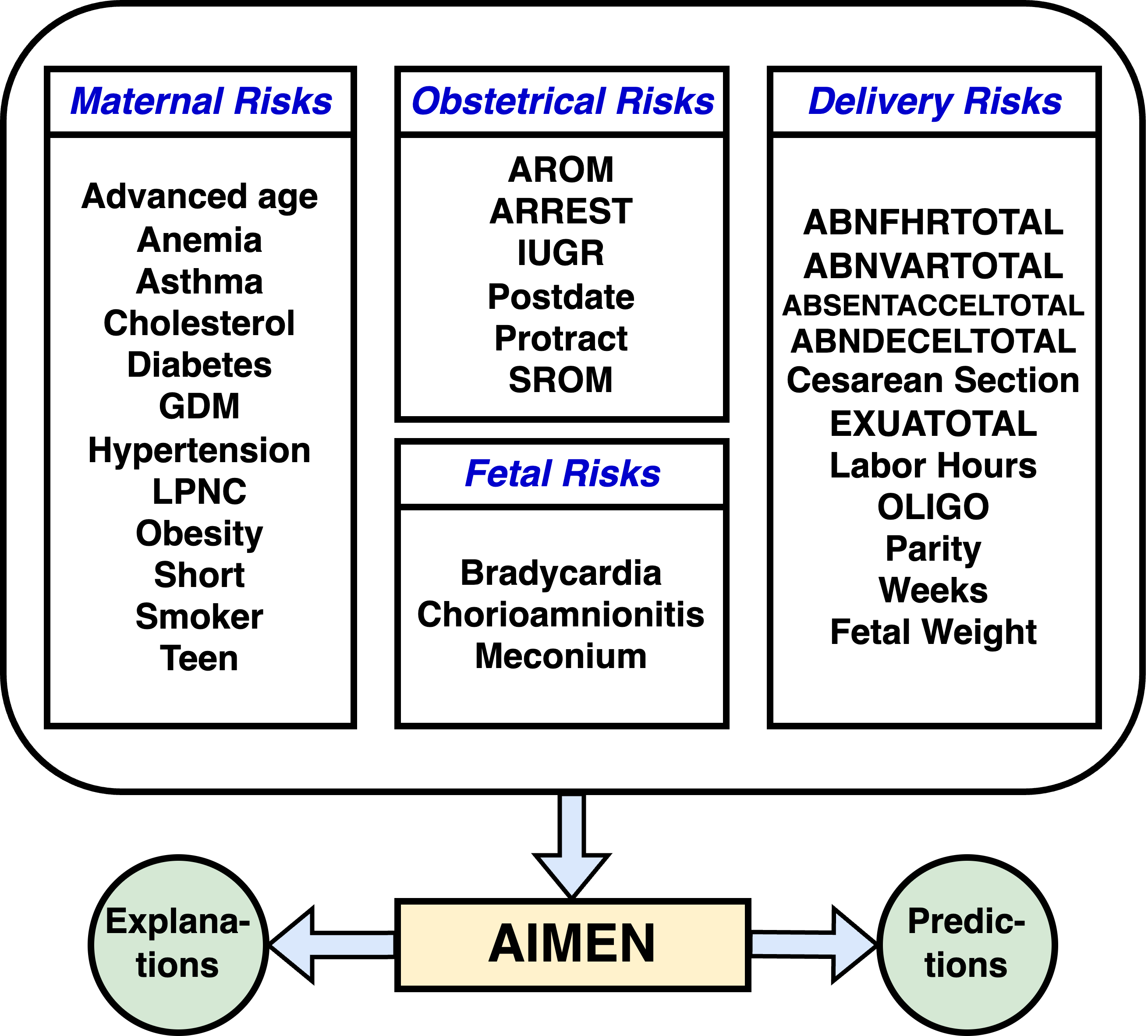}
    \caption{AIMEN uses 34 risk factors in four categories. A machine learning model is trained and used to infer the risk of cerebral palsy and other adverse delivery outcomes. It also provides counterfactual explanations of the decision made. The descriptions of the risk factors can be found in this paper \cite{mamun2023neonatal}.}
    \label{fig:fri_aimen}
\end{figure}

Electronic fetal monitoring (EFM) involves the continuous recording of fetal heart rate and the mother's uterine contractions during labor, to detect any signs of distress or abnormalities that might indicate potential complications during labor. These complications include or can lead to a large and diverse number of adverse labor outcomes, such as fetal hypoxia, acidosis, fetal distress, meconium aspiration, intrauterine growth restriction, preterm birth, neonatal encephalopathy, stillbirth, low Apgar scores, and maternal complications. Misinterpretation of EFM data is a very common allegation in malpractice litigation, claiming that such misinterpretation resulted in a lack of blood and oxygen flow to the fetal brain (birth asphyxia) \cite{michael2019cerebral}. Early signs of compromise in the neonate can be linked to a low Apgar score \cite{apgar1953proposal} or low arterial pH in the umbilical cord, and then the development of neonatal encephalopathy, a condition of altered consciousness which is suggested by many to be a requisite for cerebral palsy (CP) to have been caused by complications of labor. CP is a lifelong condition that has variable components and etiologies but functionally limits cognitive ability. One of the challenges in predicting adverse labor outcomes such as CP is the lack of standardization of definitions because CP is only one of the many possible adverse outcomes. For example, neonatal data such as arterial umbilical cord blood pH are associated with but not diagnostic of an increased risk of adverse outcomes \cite{lau2023neonatal}. Thus, numerous definitions for adverse labor outcomes have been used \cite{guedalia2021prediction, shazly2022impact}. 
Despite these known limitations, for more than 50 years, EFM has been the predominant method to evaluate fetal status and to guide clinical management. It is used in around 85\% of all labors in the United States \cite{sartwelle2018continuous, wiznitzer2017electronic}. However, it is well known that many other risk factors (RFs) are associated with adverse labor outcomes \cite{eden2018fetal, evans2021changing, evans2022earlier}. Some of these risk factors are listed in Fig. 1 and have been classified as maternal, obstetrical, fetal, and delivery risks \cite{eden2018fetal}. Thus, EFM alone does not address the relationship between risk factors and adverse labor outcomes, and a combination with other RFs has shown drastic improvements in predicting adverse labor outcomes through manual, clinical expert-derived integration in the fetal reserve index (FRI) \cite{evans2021changing}. Toward the goal of automating the integration, we describe our first steps in augmenting the clinical expert-derived FRI approach with artificial intelligence (AI) and machine learning (ML). Such a system allows updating and improving performance as more data becomes available and quantitative assessment of the weight contribution of different RFs to prediction performance. The system is intended to assist the clinicians in decision-making during labor where the large number of RFs alongside dynamic updating of risk during labor as a result of continuous EFM poses challenges in integrating these data and weighing the risks "on the fly". To this end, we propose an AI/ML-based end-to-end tool for risk analysis and explanation, AIMEN (\textbf{A}rtificial \textbf{I}ntelligence for \textbf{M}odeling and \textbf{E}xplaining \textbf{N}eonatal Health). This robust and customizable framework is designed to identify potential neonatal risks and provide reasonings for their impact on birth outcomes.

\begin{figure}[t!]
     \centering
         \includegraphics[width=0.48\textwidth]{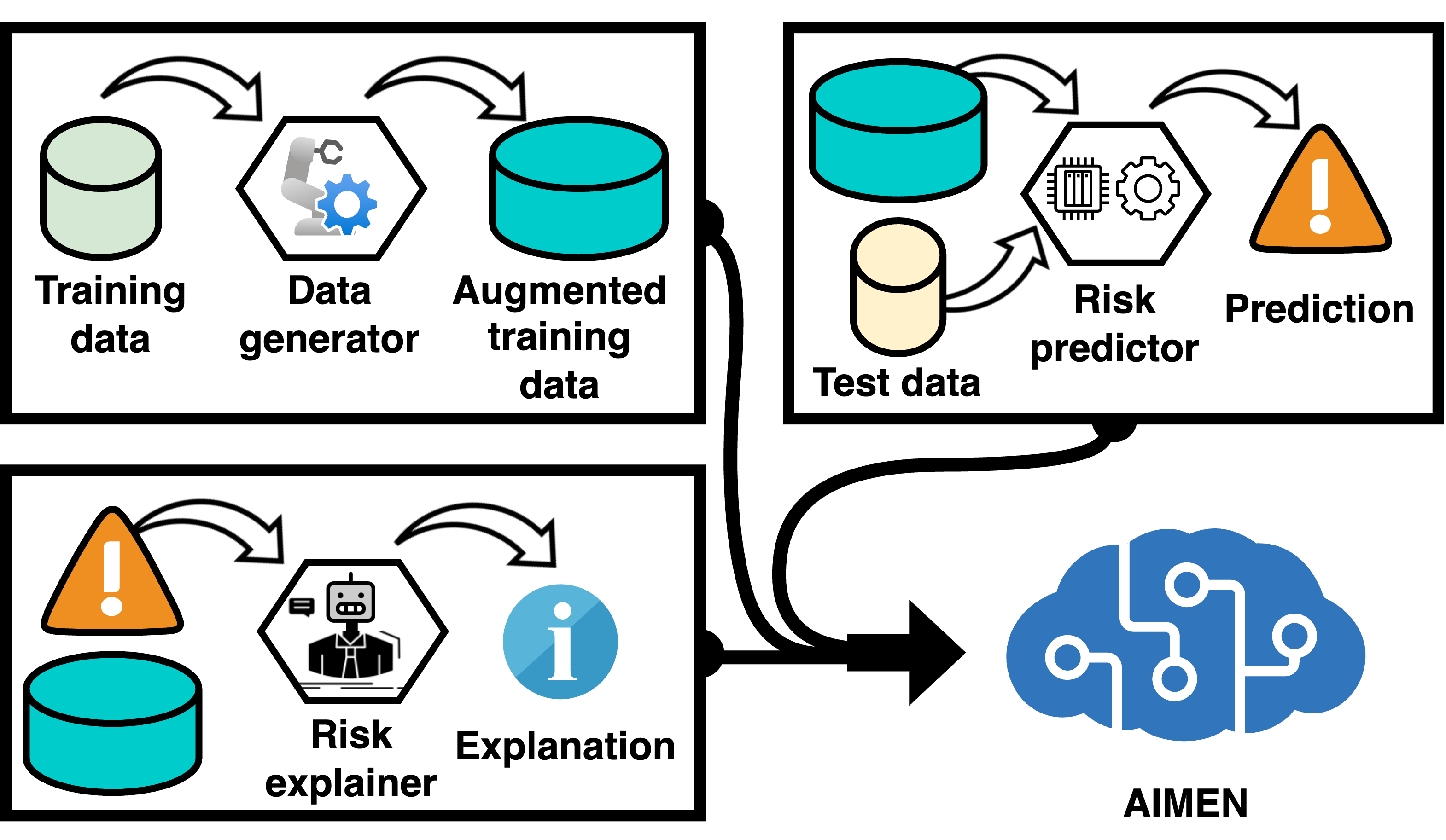}
        \caption{The AIMEN system is made of three major components: a data generator, a risk predictor, and a risk explainer. These three components allow AIMEN to learn from small and challenging datasets and provide useful explanations of a prediction or diagnosis.}
        \label{fig:block_diagram}
\end{figure}

The list of RFs used in the AIMEN system is presented in \figref{fri_aimen}. AIMEN integrates 34 different RFs in its prediction and explanation approach, including the ones used to develop and test the FRI \cite{evans2021changing}. Clinical datasets often have limitations, such as small data size, an inadequate number of samples for a specific category, or incomplete data. These challenges can make learning from these datasets difficult for most supervised learning systems. This paper addresses these challenges by providing a systematic data generation and evaluation approach. AIMEN has three major components, as shown in \figref{block_diagram}: a data generation module, a classification pipeline, and a counterfactual explanation (CE) tool that provides what-if scenarios for changing abnormal labor to normal labor.

AIMEN overcomes the challenge of small datasets by generating useful synthetic data and validating their quality. AIMEN's default data generation module is a Conditional Tabular Generative Adversarial Network (CTGAN) \cite{xu2019modeling}. We also propose two other variations of AIMEN based on the data generation method. One is the ADASYN-based framework AIMEN\_ADASYN and the other is a CTGAN-based framework with silhouette score \cite{shahapure2020cluster} restrictions called Restricted AIMEN (R-AIMEN). Silhouette score is a metric that determines the separability of different clusters. A higher overall silhouette score means that in general the samples of the same class are placed close to one another and samples from the opposite classes are placed far from one another in the hyperspace. We evaluate the quality of the generated data based on the difference between validation loss and test loss, referred to as the distribution gap.

AIMEN's downstream task is classifying abnormal labor cases with a high risk of adverse labor outcomes. Throughout this paper, unless otherwise noted, abnormal labor is defined in retrospect as a baby born with one of 3 characteristics: CP = $True$ or Apgar score at 1 minute $\leq 3$ or umbilical cord $pH \leq 7.05$. After evaluation of model parameters, we compare the performance of AIMEN system with CTGAN augmentation separately for CP, Apgar or pH criteria.   AIMEN aims to predict abnormal labor cases and thus risk of adverse labor outcomes before birth using prenatal and intrapartum RFs, by learning from a dataset where the truth is known. A dataset of 1457 such labor cases was used to develop and test the AIMEN system. 112 of the 1457 cases were abnormal and the rest were normal. Previous analysis of this dataset suggested that certain abnormal fetal heart rate (FHR) patterns are associated with a high risk of adverse labor outcomes, such as CP \cite{mamun2023neonatal}. Throughout this paper, abnormal and positive classes are equivalent terms. In the same way, normal and negative classes are equivalent.

The Explainable AI component of AIMEN provides the reasoning behind abnormal labor case classifications through CE. The explanation highlights the features that could be changed to make the prediction normal (i.e. describe what-if scenarios). These alternative situations are suggested so that minimal changes are required to the RFs to flip an abnormal class prediction to a normal class prediction. For example, for a specific abnormal case, the model can suggest that if the abnormal FHR pattern of this case were 0, while keeping everything else the same, this case would be predicted as a normal class.

To summarize, the goals of this work are: (1) formulating labor risk prediction through data generation and classification; (2) devising a method to use CE as a means to reason about the risk assessment; and (3) a method to evaluate the quality of synthetic data; and (4) conducting a comprehensive evaluation of the proposed risk assessment and counterfactual methods.
\section{Related Work}
\label{sec:related}
\subsection{AI in  Neonatal Health}
AI has affected multiple areas of health care, including obstetrics \cite{Devoe2025Current, mamun2023neonatal}, cardiovascular health \cite{zeng2014convolutional, mamun2022multimodal, mamun2022designing}, metabolic health \cite{hezarjaribi2017speech2health, arefeen2023glysim, mamun2025llm}, behavioral health \cite{saeedi2014cost, azghan2025cudle}, and medical imaging \cite{ronneberger2015u}, among many others. Ahn and Lee have published an overview of ML for obstetrics \cite{ahn2022artificial}. Davidson and Boland recently reviewed 127 distinct studies using AI/ML to improve pregnancy outcomes \cite{davidson2021towards}. However, physicians are often skeptical about AI/ML approaches in medicine in general \cite{james2022preparing}, including obstetrics \cite{sarno2023}. A recent review by Devoe et al. \cite{Devoe2025Current} analyzed 207 recent studies on artificial intelligence (AI) applications in obstetrics, highlighting rapid global growth in AI-based risk prediction, ultrasound imaging, and early screening for preterm birth, preeclampsia, and gestational diabetes. Although many models demonstrate strong predictive accuracy, few have been clinically deployed, emphasizing the need for more generalizable, validated studies and physician training to integrate AI safely into obstetric practice \cite{Devoe2025Current}. Randomized clinical trials (RCTs), the cornerstone of assessing interventions before they are incorporated in clinical practice, when applied to AI/ML-assisted interventions have solicited concerns regarding the quality of medical AI/ML RCTs \cite{plana2022randomized}. Transparency and trust as opposed to ``black box'' predictions, alongside evidence-based medicine principles and shared decision-making between patients and clinicians using AI/ML-based risk assessments, will be needed to promote their acceptance \cite{james2022preparing}. Toward this goal, we describe our first steps in developing an AI/ML approach that includes an explainable AI component, CE, to assist clinical decision-making in predicting the high risk of adverse labor outcomes, potentially increasing opportunities for interventions and mitigation. Some recent studies have incorrectly claimed computer systems have been proven to be better than expert clinical management, but all have failed to be implementable \cite{pruksanusak2022comparison, keith1995multicentre, balayla2019use, devoe2016future, devoe2000comparison}. We are therefore closely collaborating with obstetricians to increase the likelihood that the AI/ML system will be useful to them. A major challenge for developing AI/ML methods in this field is the ambiguity in definitions of gold standards and features used for model development. Neonatal data, such as arterial umbilical cord blood pH, are associated with but not diagnostic of an increased risk of adverse outcomes \cite{lau2023neonatal}. Numerous definitions for adverse or abnormal outcomes have been used \cite{guedalia2021prediction, shazly2022impact}, and more work will be needed to develop better classifiers associated with specific outcomes. This will require larger datasets, and the development of such resources is underway \cite{zhang2010contemporary}. This will also allow the application of more complex and deeper neural network models for future work. To date, such models have only been applied to EFM data, not the other RFs as features \cite{ogasawara2021deep}. As a note of caution, a recent classification of EFM data using deep learning has also indicated that more data does not always yield better results \cite{spairani2022deep}. However, we believe that there are good opportunities to enhance fetal health monitoring, especially if we combine real-time data analysis \cite{guedalia2020real} with the presentation to the clinical decision-making staff working in labor and delivery units, where the AI/ML predictions are transparent, assistive, and trustworthy \cite{devoe2016future}.

\subsection{Tabular Data Classification}
Classification with tabular data can be done with different ML algorithms. While deep learning became the default choice for computer vision and natural language processing problems, decision trees, random forests, and different ensemble methods based on decision trees and their variants are still popular choices for tabular data classification and regression. XGBoost\cite{chen2016xgboost}, TabNet \cite{arik2021tabnet}, and \danets \cite{luo2020network} are some of the recent architectures for tabular data classification. XGBoost is a scalable tree-boosting algorithm that utilizes a sequential series of decision trees where every tree corrects the mistakes of its preceding tree. This model has proven to be more accurate than deep neural networks and ensemble methods of concurrent time in multiple instances \cite{chen2016xgboost}. The self-supervised learning-based TabNet outperforms XGBoost, decision tree, and other similar methods by a significant margin on numerous datasets \cite{arik2021tabnet}. Recent studies have started exploring the potential of multilayer perceptrons (MLPs) for computer vision in terms of their performance and scalability \cite{bachmann2024scaling, liu2021pay}. In this paper, we aim to empower MLPs by supporting them with ensemble networks and an effective data augmentation methodology with generative models.

\subsection{Interpretable ML}
Interpretability is increasingly an important goal in ML research. It aims to enhance the transparency, reliability, so that patients and doctors will place more trust in an intelligent system. Molnar has provided an overview of interpretability in ML \cite{molnar2020interpretable}. Two common ways of achieving interpretability are either by making the model directly interpretable or by providing explanations of the model's decisions. The quality of an explanation is often difficult to evaluate. A position paper by Doshi-Velez and Kim \cite{doshi2017towards} makes suggestions on classifying and evaluating interpretations provided by ML models.  One specific way of achieving interpretability is by providing CEs of a particular example. For a binary classification problem, a CE of a specific prediction for an instance is a real or hypothetical scenario where some attributes of the instance would be altered to reach the opposite prediction. CEs can provide insight into what features are more likely associated with a specific outcome. They can also be used for designing interventions if they are actionable. For example, for a certain disease, the model can suggest that if the patient were 20 years younger, he or she would not face a specific outcome. However, that explanation is not actionable as a person cannot change his or her age. In contrast, a person can change their food intake patterns and thereby decrease the risk of developing insulin resistance. Accuracy, distance, and sparsity are some of the metrics that can be used when evaluating CEs. Accuracy is evaluated by whether the counterfactual example is classified as the opposite class. Distance can be measured with Euclidean distance on the normalized feature set. Finally, the sparsity is the number of features that need to be changed to convert the original outcome to a counterfactual outcome. Brughmans et al. \cite{brughmans2023nice} provide the nearest instance to CE, whereas Mothilal et al. \cite{mothilal2020explaining} use gradient descent to find optimal CE based on diversity, sparsity, actionability, and proximity.
\section{AIMEN System Design}
\subsection{Problem setup and system overview}
The goal of this paper is to estimate a classifier $f: \mathbb{R}^d \rightarrow \{0, 1\}$ that predicts an outcome variable $y \in \{0,1\}$ from a state variable $x \in \mathbb{R}^d$. The state variable $x$ is a $d$-dimensional vector of real numbers and the outcome $y$ is a boolean variable that can be either 0 or 1. In the context of neonatal risk modeling, $x$ is a vector of values of $d$ risk factors: $x_1, x_2, ..., x_d$, and the value of $y$ represents the presence or absence of a specific adverse outcome, for example, high risk of adverse labor outcome. Suppose, we have a dataset $\mathcal{D}$ with $M$ number of labor cases with their corresponding risk factors and outcomes, the state vector of the $i$-th case can be represented by $x^{(i)}$.

Estimation of the classifier, $f$, can be done in a supervised learning setting where the weights and biases can be learned by training from the data points of $\mathcal{D}$. Suppose, the test dataset is $\mathcal{D}_{test}$ where $\mathcal{D} \cap \mathcal{D}_{test} = \phi$. After training on $\mathcal{D}$, the model $f$'s performance metric on the test set $D_{test}$ is $R(f, \mathcal{D}, \mathcal{D}_{test})$. If the target performance metric is $R^{*}$, we will need to train the model on another dataset, $\mathcal{S}$, which can be a real or synthetic dataset, so that $R(f, \mathcal{D} \cup \mathcal{S}, \mathcal{D}_{test}) \geq R^{*}$. Also, $\mathcal{S} \cap \mathcal{D} = \phi$ because any common data point between these two sets is redundant and can be removed from $\mathcal{S}$ to update $\mathcal{S}$ so that the condition of disjoin is met. For the calculation of the distribution gap between the real dataset and the synthetic dataset, let us also define the loss of the classifier $f$ trained with dataset $\mathcal{D}$ and evaluated on test dataset $\mathcal{D}_{test}$ as $\mathcal{L}(f, \mathcal{D}, \mathcal{D}_{test})$.

This paper aims to solve the problem using the following steps.
\begin{enumerate}
    \item Generate synthetic dataset $\mathcal{S}$ using training dataset $\mathcal{D}$
    \item Train classifier $f$ with both $\mathcal{S}$ and $\mathcal{D}$
    \item Verify the goodness of classifier based on the performance on test data $\mathcal{D}_{test}$.
    \item Verify the goodness of $\mathcal{S}$ using distribution gap $\delta(f, \mathcal{S}, \mathcal{D}, \mathcal{D}_{test})$ given by Equations~\ref{eqn:define_delta1}, \ref{eqn:define_delta2}, and \ref{eqn:define_delta3}.
\end{enumerate}

\begin{equation}
\begin{split}
    \delta(f, \mathcal{S}, \mathcal{D}, \mathcal{D}_{test}) = \frac{\mathcal{L}_{test} - \mathcal{L}_{val}}{\mathcal{L}_{val}}
    \end{split}
    \label{eqn:define_delta1}
\end{equation}
where,
\begin{equation}
    \mathcal{L}_{test} = \mathcal{L}(f, \mathcal{D} \cup \mathcal{S}, \mathcal{D}_{test})
    \label{eqn:define_delta2}
\end{equation}
\begin{equation}
    \mathcal{L}_{val} = \text{E}_{\mathcal{D}_{val} \subset \mathcal{D} \cup \mathcal{S}}[\mathcal{L}(f, \mathcal{D} \cup \mathcal{S} - \mathcal{D}_{val}, \mathcal{D}_{val})]
    \label{eqn:define_delta3}
\end{equation}

This solution can be realized with a neonatal risk modeling system made up of an EFM to support data collection and three additional components: a data-generating tool for augmenting training data, a classifier for risk analysis, and an explainable AI component for providing counterfactual explanations of abnormal predictions. An overview of the training and evaluation pipeline is presented in \figref{block_diagram}.

\subsection{Data Collection}
Data was collected from 1462 patients. The recorded RFs include preexisting maternal conditions such as diabetes, hypertension, and cholesterol. The fetal, obstetrical, and delivery RFs and EFM data were collected. EFM features include the absence of FHR accelerations, abnormal baseline FHR, and excessive uterine activity. This dataset's full list of RFs can be found in Mamun et al. \cite{mamun2023neonatal}. A summary of some numeric features of the dataset is available in \tblref{datasummary}. Five cases were excluded because of missing RFs and the dataset was prepared with the remaining 1457 cases.

\begin{table}[htbp]
\centering
\caption{Summary of the dataset.}
\begin{tabular}{llll}
\hline
Feature      & Min & Max  & Mean $\pm$ standard deviation   \\ \hline
Maternal age (years) & 15  & 47   & 27.9 $\pm$ 5.9 \\
Gestational age (weeks)   & 27  & 42   & 38.6 $\pm$ 1.8   \\
Labor duration (hours)  & 1   & 41   & 13.4 $\pm$ 8.2   \\
Fetal weight (grams) & 950 & 4905 & 3248 $\pm$ 553  \\ \hline
\end{tabular}
\label{tbl:datasummary}
\end{table}

\subsection{Data Balancing and Augmentation}
\label{sec:balance_augment}
A major challenge with this project is the small size and imbalanced nature of the data with 1457 deliveries out of which only 112 cases were positive (abnormal). To address these issues, we increased the size of the training dataset and balanced the training dataset with the help of data generation and augmentation tools. Two different methods were used independently with additional customizable options. 

At first the training and test data were separated to prevent any data leakage. The test set was separated by choosing an equal number of real data points from the negative class (normal labor outcome) and the positive class (abnormal labor outcome). Therefore, the test was already balanced, containing only real and unseen data points. After that the training set was balanced and augmented, and the models were trained and validated using that augmented dataset. 8-fold cross-validation was used for assessing validation set performance, but the final evaluations were always on the unseen test set that only contains real data points. As a result, the test set consisted solely of real, unseen cases. This design eliminated the risk of data leakage and ensured that the reported results represent true generalization performance.

In one version of the AIMEN system (AIMEN\_ADASYN), ADASYN \cite{he2008adasyn} was used to generate synthetic data for the positive class. Then subsets of negative set data and positive set data were randomly sampled so that the size of the negative subset was lower than the size of the positive subset. Afterward, using this sampled data, negative class samples were generated. This process is repeated until the final training dataset is balanced and is at least 5 times larger than the original training dataset. It was done this way so that the final dataset had at least 5000 cases for each class, totaling at least 10000 cases in the training data.

In the other version, which is the default implementation of the AIMEN system, we employed the CTGAN \cite{xu2019modeling} model for synthetic data generation. We balanced and augmented the data in three phases: i. generated positive class data until the dataset was balanced, ii. generated negative class samples until the size was 5 times larger, and iii. generated positive class data until the dataset was balanced. This way, the final training dataset was balanced and at least 5 times the size of the original training dataset. We developed three variations of CTGAN-based augmentation based on whether any generated data was discarded. The default AIMEN system integrates all data generated by the generative model into the training dataset. In the Restricted AIMEN (R-AIMEN) variations, we used the silhouette score \cite{shahapure2020cluster} to determine whether a batch of generated data should be integrated or discarded. A batch of synthetic data was discarded if, after including this dataset with the current dataset, the silhouette score did not improve over the current silhouette score, or it did not meet a certain threshold. For example, the batch of generated data was discarded if the silhouette score after generation was not more than the previous silhouette score or not more than the predefined minimum. Note that we used the generated synthetic data only to train the models, but the final evaluation was always done on a subset of the real data.

\begin{figure}[tbh!]
    \centering
    \includegraphics[width=0.88\textwidth]{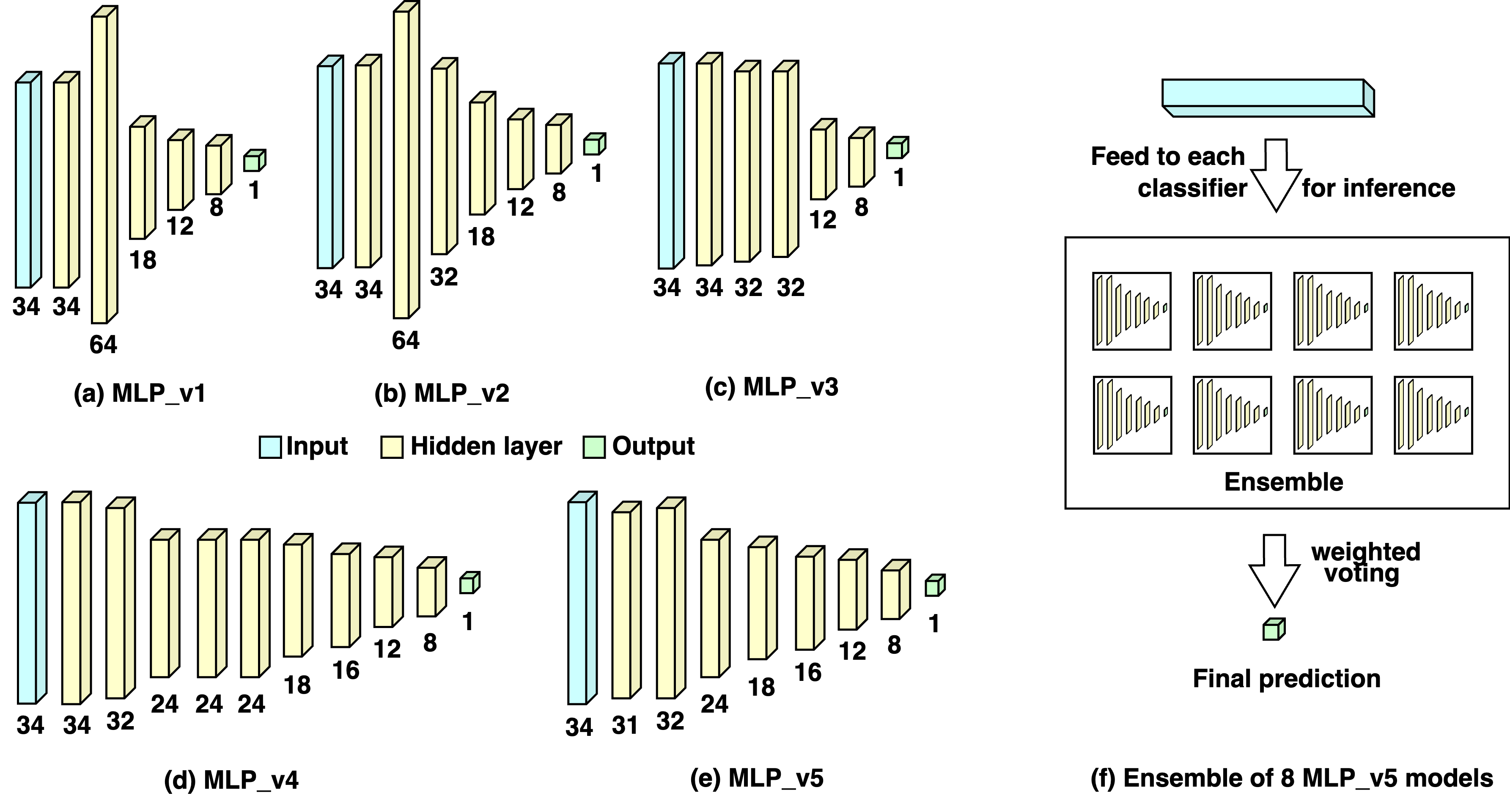}
    \caption{AIMEN's backbone is a Multilayer Perceptron (MLP) consisting of an ensemble of eight fully connected neural networks. Here, we have tried different versions of the MLP, accounting for varying width and depth as shown in the figure. Lower depth models failed to perform as well as MLP\_v5, even with more width. Moreover, MLP\_v4 also did not reach the performance of MLP\_v5, despite having higher depth, due to overfitting. Therefore, the default AIMEN was MLP\_v5. In each case, eight neural networks of the same architecture (e.g., MLP\_v5) were trained and validated on eight different folds of the cross-validation, and weighted voting among those eight models was performed through the ensemble network to classify on the test set. }
    \label{fig:architecture}
\end{figure}

\subsection{Classification}
The classification was done with ensemble learning with a group of $k$ classifier models ($k \in \mathbb{N})$ that were trained through k-fold cross-validation. Each model was trained with (k-1) folds of training data and validated on the left-out fold. Based on the performance of the validation set, some classifiers were given the voting rights to classify test data. In our experiments, the voting right was given to any classifier that achieved a macro average F1 score higher than 0.7 on its validation set. Finally, weighted voting was done among the qualified classifiers to make predictions on the test set. A classifier with a higher validation score was assigned a higher voting weight. For example, suppose the voting weights are $\alpha_1, \alpha_2, ..., \alpha_k$ for the $k$ classifier models. Now, if the prediction probabilities of a particular unseen example, $x$, by the classifiers are $f_1(x), f_2(x), ..., f_k(x) \in (0, 1)$ respectively, then the final prediction probability will be $\hat{p}= \frac{\sum_{i=1}^k \alpha_i \* f_i(x)}{\sum_{i=1}^k\alpha_i}$ and the class prediction will be $round(p)$ which returns $1$ when $p\geq0.5$, otherwise returns 0. Here, $\alpha_i = F1_i * Integer(F1_i > 0.7)$ where $F1_i$ is the macro average $F1$ score of $f_i$ on its validation set and the function $Integer(expression)$ returns 1 when $expression = True$, otherwise returns 0.


Fully-connected neural networks were employed for the classification step. Five different forms of multilayer perceptions (MLP) were tested as the backbones for AIMEN. They are named MLP\_v1 to MLP\_v5. The architecture of the MLP\_v5 model is shown in \figref{architecture}. This network has eight fully connected layers, including the output layer. The default AIMEN uses an ensemble of MLP\_v5 neural networks, but the backbone can be changed to any other option from MLP\_v1 to MLP\_v4. Based on the performance of the validation sets, weighted voting was performed among the ensemble members to calculate the output during inference.

The five MLP backbones studied (\figref{architecture}) varied in depth and width in order to identify an architecture that achieves an optimal balance between representation learning capacity and generalization. Initially, we designed a wide but shallow network (MLP\_v1) to establish a baseline with sufficient feature interactions while keeping the model simple. We then extended this idea to a wider yet slightly deeper configuration (MLP\_v2) to explore whether additional layers could enhance hierarchical feature abstraction. Next, we tested a narrower network (MLP\_v3) to assess the effect of reduced parameterization on model stability and overfitting. Building on these insights, we introduced a much deeper but progressively narrowing network (MLP\_v4), enabling the model to gradually filter and refine extracted representations by effectively learning what information to retain and what to discard across layers. Finally, we developed a comparatively shallower yet still deep model (MLP\_v5) that maintains this selective filtering ability while mitigating the risk of overfitting associated with the deeper version, such as MLP\_v4. Together, these architectural variations allowed systematic evaluation of how network width and depth influence the AIMEN system’s ability to learn clinically meaningful representations.

\subsection{Counterfactual Explanations}
One major component of the AIMEN system is its ability to highlight important features by providing alternate scenarios where an abnormal labor case could be flipped to a normal case by changing one or more RFs.  The nearest instance CEs \cite{brughmans2023nice} were calculated with our prediction module. This method considers the nearest neighbors of a specific example based on Euclidean distance after scaling the data with MinMax scaling. The nearest neighbor that belongs to the opposite class is named the nearest unlike neighbor. After the nearest unlike neighbor is found, different feature values of that neighbor are set to the value of the original example until the class label is flipped back to the original class label, in order to lower the distance and sparsity value. The nearest unlike neighbor with the feature values optimized just before the class label flipping back to the original label is chosen as the nearest instance CEs. CEs were generated for each abnormal class example from the test set to identify the major contributors to the high risk of adverse labor outcomes and potential interventions. 

\subsection{Performance Metrics}
As the dataset was highly imbalanced, it was important to evaluate a classifier's performance with multiple metrics besides accuracy. The performance metrics reported are accuracy, sensitivity, specificity, positive class F1 score, negative class F1 score, average F1 score, and area under the receiver operating characteristic curve (AUROC\footnote{The formal definition of AUROC can be found in \cite{bradley1997use}.}). They are described below. Suppose, there are $\mid \mathcal{D}_{test} \mid$ = $M$ test examples and the symbols $TP$, $TN$, $FP$, $FN$ represent the numbers of true positive, true negative, false positive, and false negative predictions respectively. Also, suppose, $y^{(i)}$ is the true label (0 or 1) of $i$-th test example and $p^{(i)}$ is the predicted probability with which the $i$-th test example belongs to class 1. Then the evaluation metrics can be calculated as:

\begin{itemize}
    \item Binary cross entropy loss:  $\frac{1}{M} \sum_{i=1}^M{(y^{(i)}\log(p^{(i)}) + (1 - y^{(i)})\log(1 - p^{(i)}))}$,
    \item Sensitivity: $\frac{TP}{TP+FN}$.
    \item Specificity: $\frac{TN}{TN+FP}$.
    \item Positive predictive value (PPV): $\frac{TP}{TP+FP}$.
    \item Negative predictive value (NPV): $\frac{TN}{TN+FN}$.
    \item F1 score for positive class ($F_1^+$): $\frac{2 \times PPV \times Sensitivity}{PPV + Sensitivity}$
    \item F1 score for negative class ($F_1^-$): $\frac{2 \times NPV \times Specificity}{NPV + Specificity}$
    \item Average F1 score ($F_1$): $(F_1^+ + F_1^-)/2$
\end{itemize}

F1 scores for both classes are presented in the results tables and figures, along with the macro average F1 score to provide an idea of how the models are doing for the positive class examples and the negative class examples.
\section{Results}

\label{sec:results}
We compared several different backbones of AIMEN and investigated different choices of parameters to find the optimal configuration for prediction and CEs. Although synthetic data was generated to augment AIMEN, all test set results reported in this study were derived from evaluations conducted exclusively on real, previously unseen data.

\subsection{CTGAN vs ADASYN}
Synthetic data generation was employed with both CTGAN and ADASYN, and overall, CTGAN-generated data were more helpful for the downstream task. In \figref{float_bars}, we compare different methods of data generation. Data generation with CTGAN allows specifying the categorical variables, and the generated values for those variables will be integer values of 0 and 1. For the numerical variables, however, by default, CTGAN generates data that is out of the range seen in training data. For example, labor hours are present in the training data with only integer values $\geq$ 0. But CTGAN also generated examples with negative values. We conducted multiple rounds of experiments where i) we allowed those negative values to be used for training the classifiers, or ii) we replaced any negative value with 0, as a negative value for a duration may not make sense. Allowing negative values in the generated data for training the models made the downstream task more accurate, as shown in \figref{float_bars}a. This may be because labor starts before a mother comes to the hospital.

\subsection{Validation of the Synthetic Data}
\begin{figure}[htbp]
    \centering
        \includegraphics[width=0.50\textwidth]{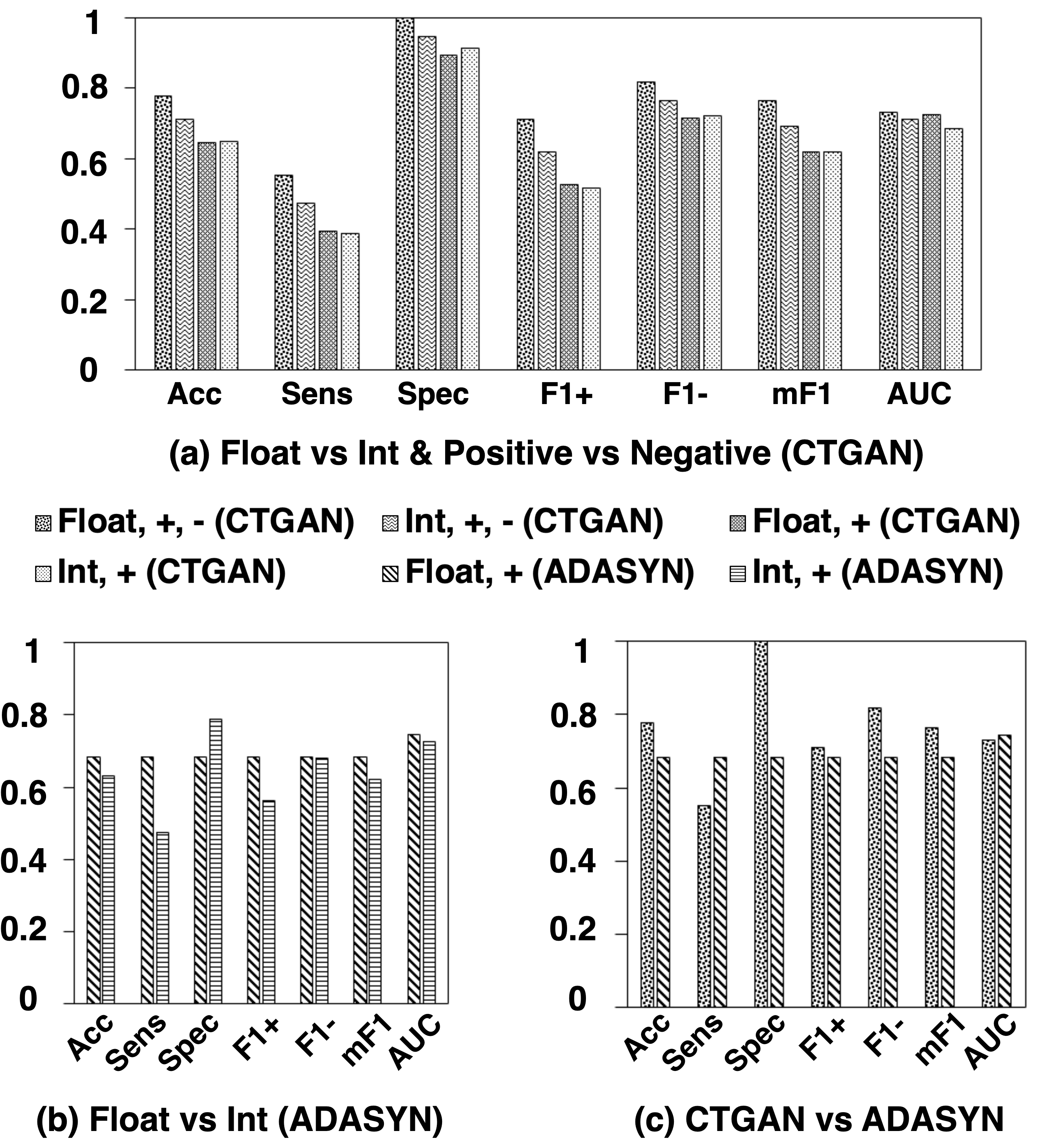}
    
    \caption{Performance metrics on the test set using different methods of data generation. The real training data features have only positive integers; however, generated data can have fractional and out-of-range (negative) values by default. Float means the classifiers were trained with fractional values, and Int means the generated data were converted to integers before training the classifiers. F1+ is the F1 score of the abnormal class, F1- is the F1 score of the normal class, and mF1 is the macro average F1 score. +, - in the legend represents both positive and negative values were present in the generated training data, whereas + means all the values of the training data were positive. We experimented training our models for both cases: when negative feature values in the synthetic dataset are rounded up to zero, and the cases when they are kept as-is. The motivation was to investigate whether not rounding up feature values can allow for exploration of more precise decision boundaries, i.e., better results on the unseen real test set. Using a modest fraction of off-support synthetic examples as a boundary regularizer is supported by prior work in adversarial/virtual-adversarial training, outlier exposure, vicinal data augmentation (mixup), and domain randomization \cite{goodfellow2014explaining, zhang2017mixup, miyato2018virtual}. These methods all show that carefully chosen synthetic or out-of-distribution samples can improve decision-boundary behavior and generalization. In our experiments, we found that retaining small negative synthetic values (rather than rounding them to zero) improved held-out performance, consistent with this literature \cite{goodfellow2014explaining, zhang2017mixup, miyato2018virtual}.}
    \label{fig:float_bars}
\end{figure}

To validate the use of the synthetic data, we assessed the similarity between real and augmented data distributions to investigate the extent to which the synthetic data captures clinical realism. We intentionally did not restrict CTGAN, e.g. we permitted CTGAN to generate slightly out-of-range values, which we consider a form of implicit regularization. In adversarial and virtual adversarial training, models are intentionally exposed to off-manifold or unrealistic perturbations to enhance generalization and refine decision boundaries~\cite{goodfellow2014explaining, miyato2018virtual, zhang2017mixup}. Although such perturbations may not be clinically plausible, they have been shown to yield more robust and stable models. Following this intuition, we experimented with two pre-processing strategies: (\emph{i}) rounding negative feature values in the synthetic dataset to zero, and (\emph{ii}) retaining them as-is. Allowing a small fraction of out-of-range samples (e.g., negative labor hours) effectively introduces “boundary-challenging” examples that encourage smoother and more stable decision surfaces. Furthermore, we hypothesize that filtering out-of-bound values to the nearest domain-accepted value can disrupt intrinsic feature correlations within each synthetic datapoint, reducing internal coherence. Importantly, this mechanism was applied only to training data in our experiments, while all test data consisted solely of real samples, ensuring that clinical realism in evaluation was preserved. Moreover, this property of AIMEN can be easily disabled to constrain synthetic feature values within domain boundaries, and we have reported results for that configuration as well in \figref{float_bars}a. Thus, our approach provides an additional exploratory avenue to enhance model robustness without compromising clinical validity or interpretability. Our ablation results indicate that preserving these small negative values leads to improved performance on unseen test data, as seen in \figref{float_bars}a, consistent with prior findings that controlled off-manifold augmentations can serve as effective regularizers~\cite{goodfellow2014explaining, miyato2018virtual, zhang2017mixup}.
To further assess the similarity between real and augmented data distributions, both Maximum Mean Discrepancy (MMD) and Wasserstein Distance \cite{panaretos2019statistical} were computed across five independent experiments and the results are listed in \tblref{mmd_wd}. These metrics quantify the alignment between the original training data, CTGAN-augmented data, and the test set. Lower values indicate a smaller distributional gap and higher similarity. 

Before computing these metrics, all feature sets were normalized using min–max normalization, where the minimum and maximum feature values were derived from the reference training features of the first experiment. This ensured a consistent feature scale across all runs. 

The MMD was calculated using an RBF kernel, with the kernel bandwidth parameter $\gamma$ determined automatically through the median heuristic, which computes the median of pairwise Euclidean distances between samples of the two datasets. For any pair of datasets X and Y, the kernel matrices $K_{XX}$, $K_{YY}$, and $K_{XY}$ were then used to estimate the empirical squared MMD value, followed by taking the square root to obtain the final score. A lower MMD value indicates that the two distributions are more similar in the reproducing kernel Hilbert space (RKHS) \cite{berlinet2011reproducing}.

The Wasserstein distance was computed using the sliced Wasserstein approach, which is efficient for high-dimensional feature spaces. For each comparison, 100 random projection vectors were sampled, and the one-dimensional Wasserstein distance was calculated between the projected samples using the \texttt{scipy.stats.wasserstein\_distance} function. The final distance was obtained by averaging across all projections. This approach approximates the Earth Mover’s Distance while maintaining computational efficiency.

The Wasserstein distance between the original training and CTGAN-augmented data is 0.085, slightly higher than 0.075 observed between the training and test data, indicating that the CTGAN-augmented distribution remains close to, but slightly more diverse than, the original training distribution. Similarly, the MMD values follow the same trend, showing that the augmented data moderately differ from the training set while maintaining proximity to the test distribution. These values together suggest that CTGAN augmentation introduces realistic variability without deviating substantially from the real data manifold.

\begin{table}[h!]
\centering
\caption{Distributional similarity metrics using sliced Wasserstein distance and MMD with RBF kernel. In this table, the training set includes only real training data, the test set includes only unseen real test data, and the augmented training set combines real training samples as well as synthetic data generated near the training distribution. Lower values indicate higher similarity. The results show that CTGAN augmentation adds meaningful variability while maintaining close alignment with the real training distribution.}
\label{tbl:mmd_wd}
\begin{tabular}{lcc}
\hline
\textbf{Comparison} & \textbf{MMD (RBF Kernel)} & \textbf{Wasserstein Distance (Sliced)} \\
\hline
Train vs Train CTGAN Augmented & 0.188 & 0.085 \\
Train vs Test & 0.129 & 0.075 \\
Train CTGAN Augmented vs Test  & 0.121 & 0.080 \\
\hline
\end{tabular}
\end{table}

\subsection{Performance on the Training and Validation Sets}
\begin{figure}[htbp]
    \centering
        \includegraphics[width=0.50\textwidth]{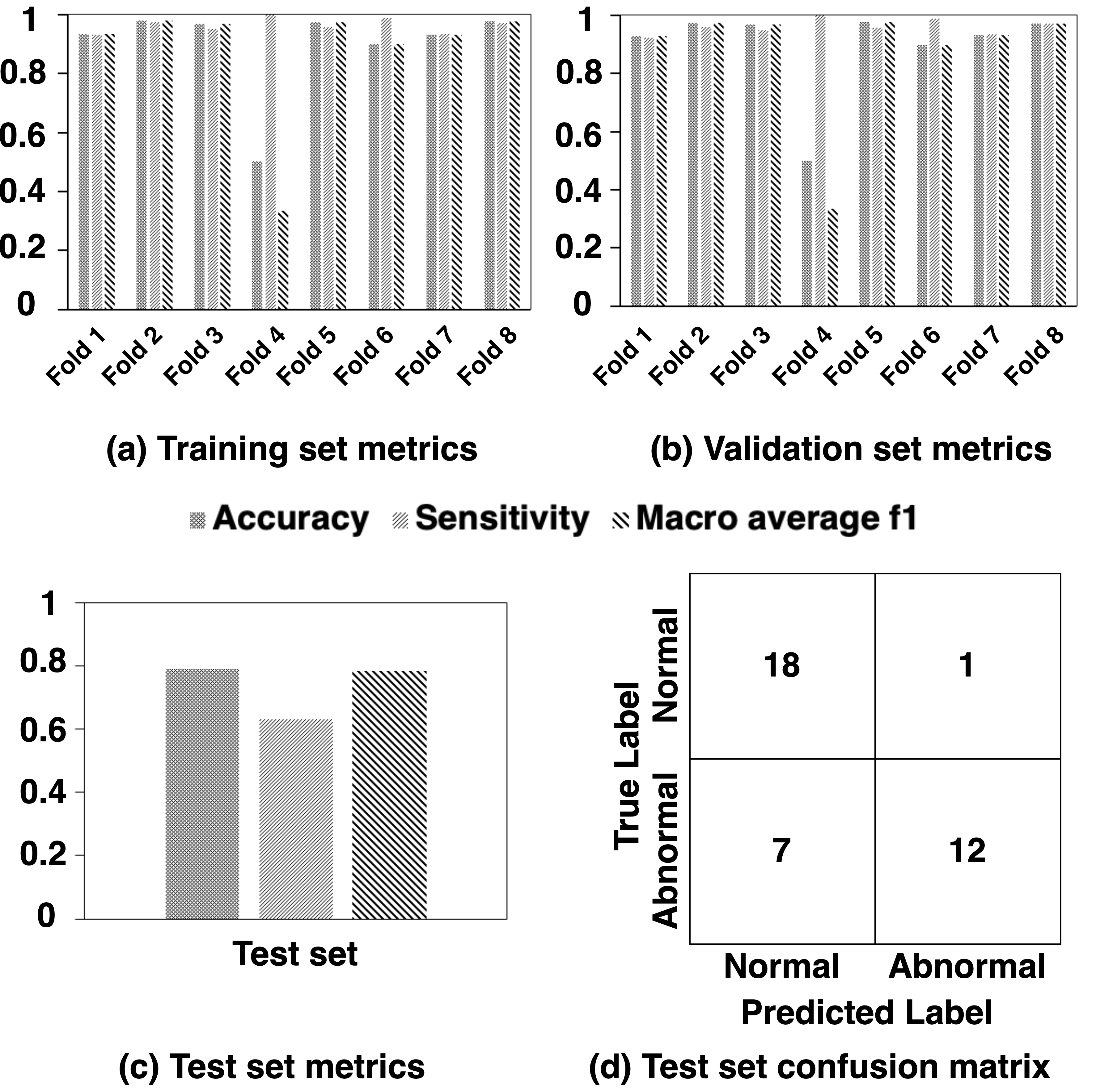}
    
    \caption{Training, validation, and test set metrics along with the test set confusion matrix with the AIMEN system with CTGAN data augmentation tool and MLP v5 backbone. (a) Training set metrics. (b) It can be seen that Fold 4 has a validation set F1 score less than 0.7, so the models trained for that fold will not be included in the final ensemble network of AIMEN that is used to predict on the test set. (c) AIMEN's performance on the test set is evaluated on real and unseen data points. (d) Confusion matrix on the test set.}
    \label{fig:all_bars}
\end{figure}

When a model performs well on both the training and validation metrics, it indicates that it can learn representations and generalize, which prepares it well for unseen data. In \figref{all_bars}, we can see that our model achieves macro average F1 scores over 0.9 in most of the training and validation set experiments. This finding indicates that the model is not underfitting or overfitting.

One challenge of these experiments is that the training and validation sets have both real and synthetic data. So, if the synthetic data does not represent the distribution of the real data, the performance on the validation set may not equate to the performance on the test set. We therefore tested the results when all the examples were from real data in \figref{all_bars}. We can see that the model achieved an accuracy of 0.789, a sensitivity of 0.632, and a macro average F1 score of 0.784 on a balanced test set of real data. One challenge of evaluating the method on the test set was the small data size. As we had only 112 abnormal class examples in the whole dataset and a large part of them were used in training and data generation, we had to exclude them from the test set. The test set had 38 examples: 19 normal and 19 abnormal. The confusion matrix of \figref{all_bars} shows that the model identified 12 of the 19 positive class examples, corresponding to a sensitivity of 0.632, while identifying 18 out of 19 negative class examples, corresponding to a specificity of 0.947.

Finally, voting rights are given to only these classifiers with a macro average F1 score $>$ 0.7 on the corresponding validation set. In the case of \figref{all_bars}, the model for Fold 4 was therefore excluded from voting when evaluating the test set.

\subsection{Different Backbones of AIMEN}
Five different neural networks were tested as the backbone of the AIMEN system (\figref{architecture}. The summary of their performance is presented in \tblref{backbone_chooser}.  Each backbone was trained and tested five times on different training and test sets, and the average performance was reported. The default AIMEN (uses MLP\_v5 backbone) achieves the best result on all performance metrics, as reported in \tblref{backbone_chooser}.

In total, five different MLP backbones were explored to optimize performance on a tabular dataset with 34 features. The architectures ranged from shallow (MLP\_v1) to deeper and more balanced structures (MLP\_v5). The superior performance of MLP\_v5 can be attributed to its wider input layer, which effectively captures the full feature space, and its progressive reduction in layer size, which supports hierarchical feature refinement while reducing overfitting. The ensemble of eight MLP\_v5 models, trained across folds and combined via weighted voting, further enhanced generalization and contributed to AIMEN’s robust performance on unseen data.

\begin{figure}
    \centering
    \includegraphics[width=0.75\linewidth]{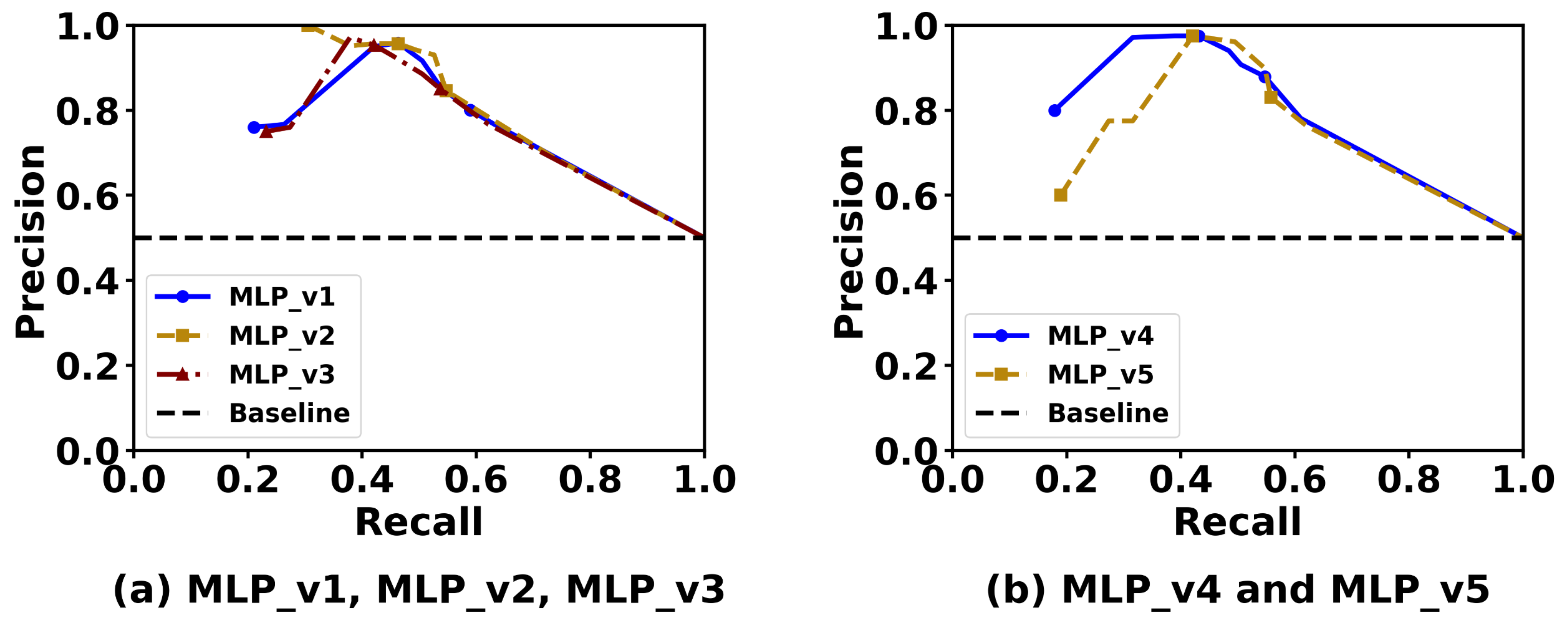}
    \caption{Precision-recall curves of the AIMEN system for all different backbones.}
    \label{fig:auprc_curves}
\end{figure}

Although MLP\_v5 achieved the best accuracy, sensitivity, F1 score, and AUROC metrics, MLP\_v4 outperformed other baselines in the area under the precision-recall curve (AUPRC) metric. Precision-recall curves for the AIMEN system for the different backbones are compared in \figref{auprc_curves}.

\begin{table}[htbp]
\centering
\caption{Comparison of the performance of the downstream classification task using different backbones of the AIMEN system. In this experiment, AIMEN with MLP\_v5 backbone achieved the best performance in accuracy, sensitivity, specificity, F1 score, and AUROC. Here, AUPRC is the area under the precision-recall curve. Boldfaced values represent the best performance values in the corresponding column.}
\begin{tabular}{llllllllllll}
\hline
\textbf{Backbone} & \textbf{Acc} & \textbf{Sens} & \textbf{Spec} & \textbf{PPV} & \textbf{NPV} &  \textbf{F1+} & \textbf{F1-} & \textbf{Avg F1} & \textbf{AUROC} & \textbf{AUPRC}\\
\hline
MLP\_v1          & 0.726        & 0.474         & 0.979  &  0.814 & \textbf{0.572}      & 0.631        & 0.782        & 0.706           & 0.755  & 0.593        \\
MLP\_v2          & 0.726        & 0.474         & 0.979   &   \textbf{0.883} & 0.570    & 0.631        & 0.782        & 0.706           & 0.748  & 0.532        \\
MLP\_v3          & 0.721        & 0.463         & 0.979    &   0.822 & 0.565  & 0.622        & 0.779        & 0.700           & 0.744  & 0.577        \\
MLP\_v4          & 0.732        & 0.484         & 0.979    &  0.861 & 0.570    & 0.640        & 0.785        & 0.713           & 0.759 & \textbf{0.643 }         \\
MLP\_v5          & \textbf{0.753}        & \textbf{0.516}     & \textbf{0.989} &  0.785 & 0.570     & \textbf{0.674}        & \textbf{0.800}  &  \textbf{0.737}           & \textbf{0.759}    & 0.602     \\
\hline
\end{tabular}
\label{tbl:backbone_chooser}
\end{table}

\subsection{Performance of the Baseline Models vs AIMEN}

\tblref{baseline_vs_aimen_combined} presents the mean accuracy and F1 scores of the baseline models compared to the AIMEN framework. Among all models evaluated, AIMEN achieved the highest performance with an average accuracy of 0.753 and F1 score of 0.737, outperforming all other baselines across both metrics. XGBoost and LightGBM showed competitive results, yet their confidence intervals suggest less consistent performance compared to AIMEN. 

The AIMEN backbones MLP\_v1 to MLP\_v5 without the AIMEN wrapper were also evaluated as baselines, and they exhibited moderate performance, with the best among them (MLP\_v5) reaching an accuracy of 0.700 and an F1 score of 0.666. TabNet lagged significantly behind all models, reflecting the difficulty of capturing task-specific features through its sparse attention-based learning in this dataset.

Overall, AIMEN demonstrates superior predictive capability, likely due to its multimodal integration and explainability-driven design, which enable more robust representation learning across heterogeneous data sources. A more detailed discussion of the results with statistical tests and confidence intervals is discussed in \secref{statistical_analysis}.

\begin{table}[h!]
\centering
\caption{Performance comparison between baseline models and AIMEN, including mean $\pm$ standard deviation and 95\% confidence intervals for accuracy and macro average F1 scores. Boldfaced values represent the best results for the corresponding metric or confidence interval bound. Statistical tests are discussed in detail in \secref{statistical_analysis}.}

\begin{tabular}{lcccc}
\hline
\textbf{Model} & \textbf{Accuracy} & \textbf{F1 Score} & \textbf{Accuracy 95\% CI} & \textbf{F1 95\% CI} \\
\hline
XGBoost & 0.742 $\pm$ 0.043 & 0.726 $\pm$ 0.051 & [0.688, 0.796] & [0.663, 0.789] \\
LightGBM & 0.726 $\pm$ 0.076 & 0.700 $\pm$ 0.097 & [0.632, 0.821] & [0.579, 0.821] \\
TabNet & 0.532 $\pm$ 0.043 & 0.430 $\pm$ 0.091 & [0.478, 0.585] & [0.318, 0.543] \\
Standalone MLP\_v1 & 0.653 $\pm$ 0.104 & 0.590 $\pm$ 0.163 & [0.523, 0.782] & [0.387, 0.792] \\
Standalone MLP\_v2 & 0.668 $\pm$ 0.100 & 0.614 $\pm$ 0.151 & [0.545, 0.792] & [0.427, 0.802] \\
Standalone MLP\_v3 & 0.647 $\pm$ 0.096 & 0.584 $\pm$ 0.154 & [0.528, 0.767] & [0.392, 0.775] \\
Standalone MLP\_v4 & 0.684 $\pm$ 0.113 & 0.636 $\pm$ 0.156 & [0.544, \textbf{0.825}] & [0.442, \textbf{0.830}] \\
Standalone MLP\_v5 & 0.700 $\pm$ 0.063 & 0.666 $\pm$ 0.086 & [0.621, 0.779] & [0.560, 0.772] \\
\danets & 0.695 $\pm$ 0.063 & 0.666 $\pm$ 0.082 & [0.616, 0.774] & [0.564, 0.768] \\
AIMEN & \textbf{0.753 $\pm$ 0.044} & \textbf{0.737 $\pm$ 0.050} & [\textbf{0.698}, 0.807] & [\textbf{0.675}, 0.800] \\
\hline
\end{tabular}
\label{tbl:baseline_vs_aimen_combined}
\end{table}

\subsection{Effect of Decision Threshold}
\begin{figure}[htbp]
    \centering
        \includegraphics[width=0.55\textwidth]{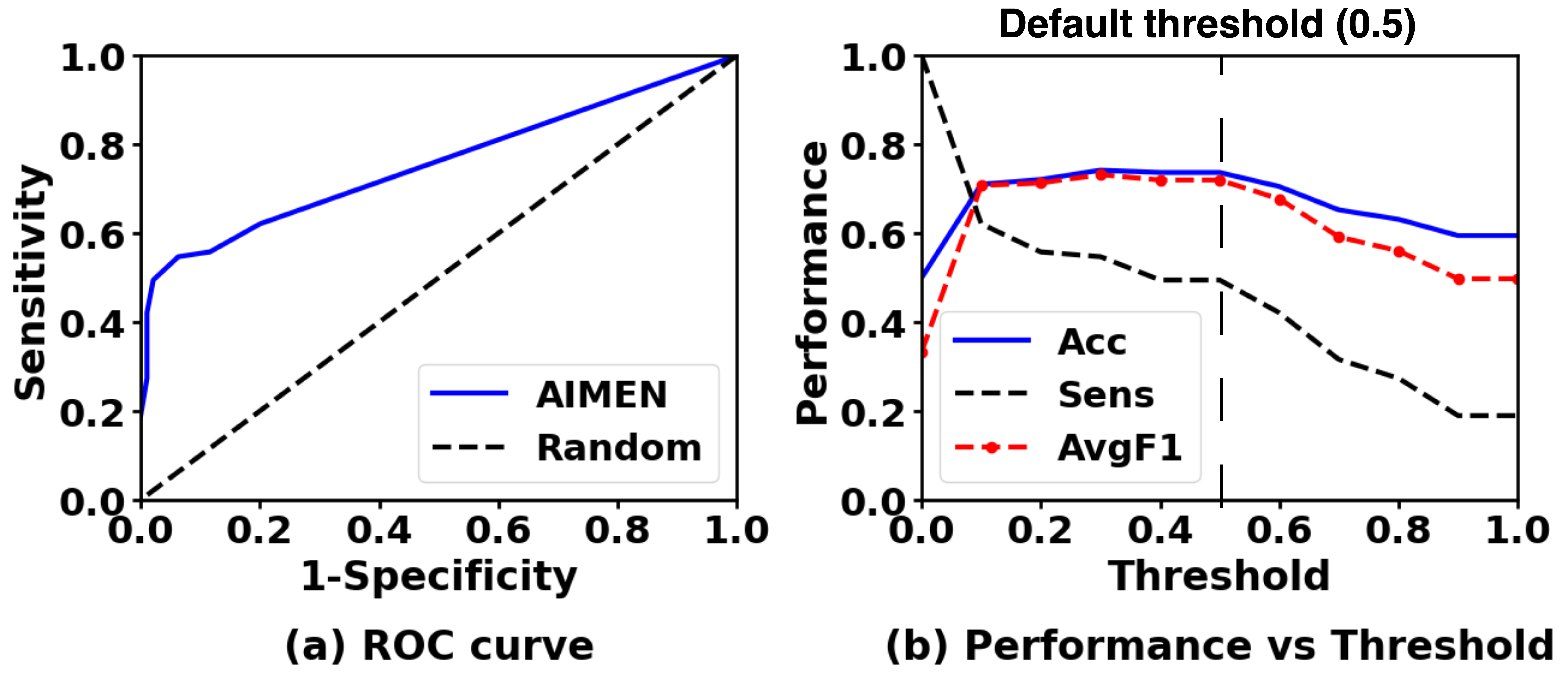}
    
    \caption{ROC curve and the performance of the classification based on the decision threshold. (a) The ROC curve using true positive rate (sensitivity) vs false positive rate (1 - specificity) for AIMEN with MLP\_v5 backbone is shown against the ROC of a random classifier. (b) Accuracy, sensitivity, and average F1 score are presented for different thresholds.}
    \label{fig:roc_combined}
\end{figure}

The default decision threshold chosen throughout this paper is 0.5, meaning, the output probability of the classifier is $\geq$ 0.5, a case is classified as abnormal, otherwise, normal. In \tblref{backbone_chooser}, we can see that the AIMEN v5 system has a sensitivity of 0.516 when the decision threshold is 0.5. To check how the system's performance changes with different decision thresholds, we plot the receiver operating characteristic (ROC) curve and the classification performance of the system in \figref{roc_combined}. From this figure, the physicians can decide which decision threshold is suitable for labeling an example as abnormal.

\subsection{Evaluating the Counterfactuals}
\begin{figure}[h!]
     \centering

     \includegraphics[width=0.86\textwidth]{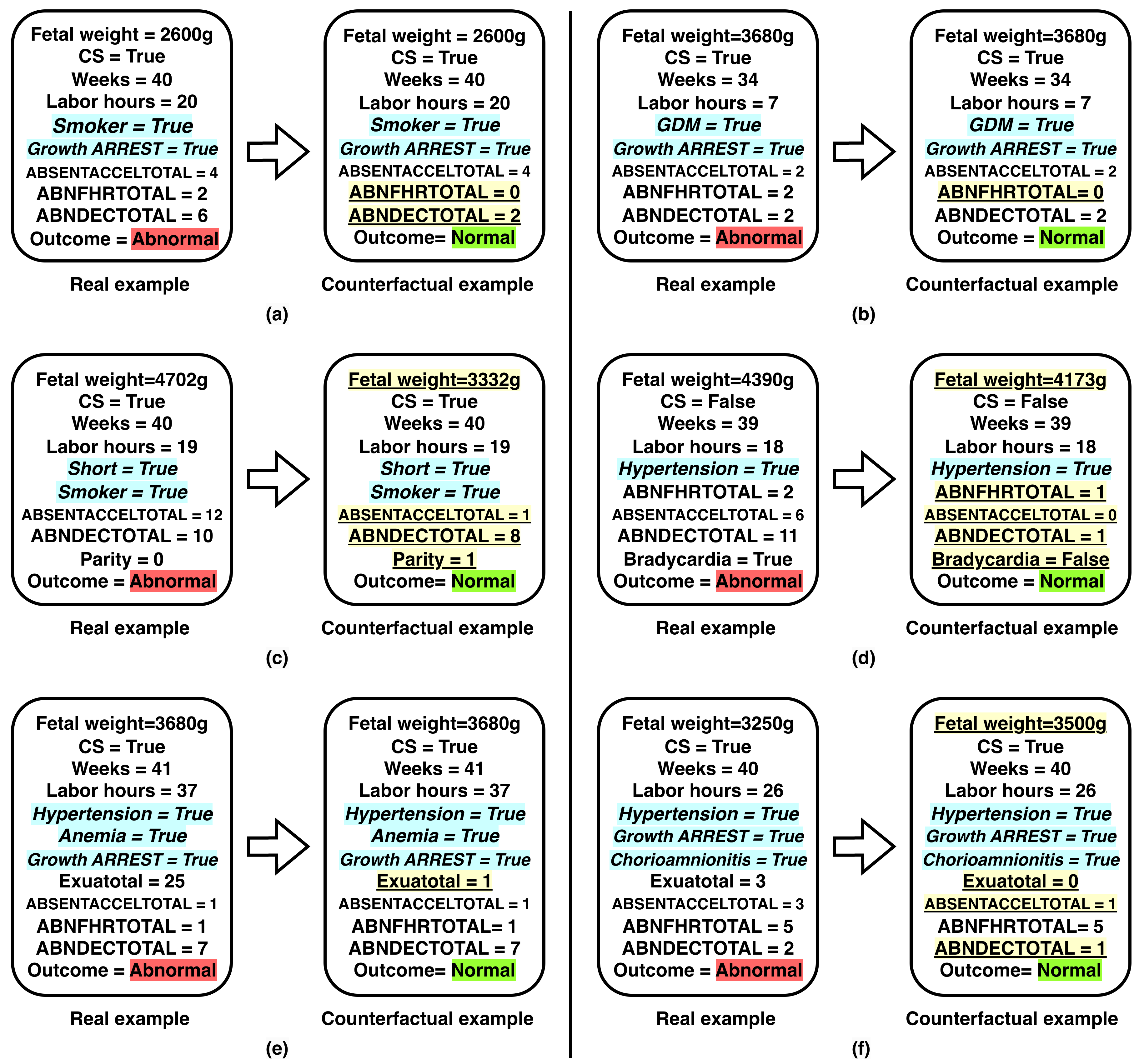}
        \caption{Examples of CE using the nearest instance CE algorithm. Attributes to be changed are underlined. Here, a specific example of an abnormal class is presented with its corresponding CE. The values of all the unchanged attributes are not shown here for clarity. However, existing maternal risk factors and the presence of growth arrest of the fetus or chorioamnionitis are mentioned and highlighted in blue when applicable. In many cases, fetal growth arrest was present, defined as fetal weight below the 10th percentile for gestational age. In case (a), changing two attributes (abnormal FHR and abnormal decelerations) would classify the example as a normal class example. For case (b), changing only the abnormal fetal heart rate would classify the example as a normal outcome. In case (c), the mother has short stature ($<$5'1'') and a smoking habit; a 4.7 kg fetus further increases the risk of complications \cite{beta2019maternal, chatfield2001acog, yearwood2023association}. The CE indicates that reducing fetal weight, having prior pregnancy experience (parity = 1 instead of 0), and lowering the absence of fetal heart rate acceleration and deceleration could minimize the risk of abnormal outcome. In case (d), preventing bradycardia, reducing abnormal fetal heart rate deceleration and overall abnormal fetal heart rate, along with a slight reduction in fetal weight, represents the minimal change needed to avoid an abnormal labor outcome. In case (e), where the mother has hypertension and anemia, excessive uterine contractions must be substantially reduced to ensure a safe labor outcome. In case (f), with hypertension and chorioamnionitis, the minimal change required is maintaining a normal number of uterine contractions and lowering the absence of fetal heart rate acceleration and deceleration to prevent an abnormal labor outcome. These CEs suggest that an abnormal fetal heart rate and a high fetal weight (when close to or above 4500 g) may increase the risk of abnormal delivery, and that avoiding such occurrences could help prevent it, which aligns with current clinical understanding \cite{nihWhatRisk, beta2019maternal, chatfield2001acog}. It suggests that the AIMEN system is able to determine the true risk factors behind the abnormal labor outcomes. Definitions of the risk factors are provided in this paper by Mamun et al. \cite{mamun2023neonatal}. Future work will address the actionability of the CEs for proper intervention design.
}
        \label{fig:counterfactual_demo}
\end{figure}

We present six different counterfactual examples produced by our methods in \figref{counterfactual_demo}.  In the case of \figref{counterfactual_demo}a, modifying two attributes: abnormal fetal heart rate (ABNFHRTOTAL) and abnormal decelerations (ABNDECTOTAL) changes the prediction from an abnormal to a normal delivery outcome. In case of \figref{counterfactual_demo}b, altering only the abnormal fetal heart rate value (ABNFHRTOTAL) is sufficient to achieve the same effect. These examples assist clinicians in decision-making by pinpointing the specific factors that trigger risk alerts and contribute most strongly to adverse labor outcomes. In \figref{counterfactual_demo}c, the mother has short stature ($<5'1''$) and a smoking habit, while a 4.7\,kg fetus further elevates the risk of delivery complications \cite{beta2019maternal, chatfield2001acog, yearwood2023association}. The corresponding counterfactual explanation (CE) suggests that reducing fetal weight, having prior pregnancy experience (parity $=1$ instead of $0$), and lowering the absence of fetal heart rate acceleration and deceleration could minimize the risk of an abnormal outcome. In \figref{counterfactual_demo}d, preventing bradycardia, reducing abnormal fetal heart rate deceleration and overall abnormal fetal heart rate, along with a slight reduction in fetal weight, represents the minimal intervention required to shift the prediction toward a normal labor outcome.

In case of \figref{counterfactual_demo}e, where the mother exhibits both hypertension and anemia, the CE reveals that a substantial reduction in excessive uterine contractions is required to achieve a favorable labor outcome. Likewise, in \figref{counterfactual_demo}f, which involves maternal hypertension and chorioamnionitis, stabilizing uterine contractions within a normal range and decreasing the absence of fetal heart rate acceleration and deceleration constitute the minimal modifications necessary to avert an abnormal delivery. Taken together, these CEs indicate that aberrant fetal heart rate dynamics and elevated fetal weight, particularly when nearing or exceeding 4500\,g, significantly heighten the likelihood of adverse labor outcomes. Mitigating these factors may therefore reduce delivery-related risks, consistent with established clinical findings \cite{nihWhatRisk, beta2019maternal, chatfield2001acog}. These results underscore the ability of the AIMEN system to uncover clinically relevant and physiologically interpretable determinants of abnormal labor outcomes. While the current implementation does not account for the actionability of the usually immutable features (such as fetal weight), it focuses on what-if scenarios with minimal changes that can provide insights for clinicians.

The CEs are evaluated based on the average distance and the average sparsity in our experiments. The average distance is the average Euclidean distance between the normalized real example and the corresponding normalized counterfactual example pairs. The average sparsity is the average number of variables that need to be changed to flip the prediction from the abnormal class to the normal class.  We present a summary of this evaluation in \tblref{evaluate_counterfactuals}. The feature dimension of the dataset is 34. An average distance of 0.33 and an average sparsity of 2.50 means that with this method, on average, a CE is located 0.33 units away from a real example in the 34-dimensional hyperspace, and on average 2.5 out of the 34 attributes need to be changed for an abnormal class example to convert to a normal class example.

\begin{table}[htbp]
\caption{Evaluation of the CEs after using different architectures. In this case, the dimension of the feature vectors is 34, which should be considered while interpreting the results.}
\centering
\begin{tabular}{lllll}
\hline
Model  & Backbone & Accuracy & Distance & Sparsity \\ \hline
AIMEN & MLP\_v1 & $1.00$ & $0.27 \pm 0.10$ & $2.83 \pm 1.40$ \\
AIMEN & MLP\_v2 & $1.00$ & $0.26 \pm 0.12$ & $2.73 \pm 1.54$ \\
AIMEN & MLP\_v3 &  $1.00$ & $0.34 \pm 0.28$ & $2.33 \pm 1.41$ \\
AIMEN & MLP\_v4 & $0.91$ & $0.34 \pm 0.26$ & $2.64 \pm 1.55$ \\
AIMEN & MLP\_v5 &  $1.00$ & $0.33 \pm 0.27$ & $2.50 \pm 1.36$ \\
AIMEN\_ADASYN & MLP\_v5 & $0.68$ & $0.33 \pm 0.28$ & $1.90 \pm 1.48$ \\
\hline
\end{tabular}
\label{tbl:evaluate_counterfactuals}
\end{table}

\tblref{feature_changes}, highlights the top 10 features that are frequently recommended to be changed in the counterfactuals are provided. It can be seen that the absence of fetal heart rate acceleration (ABSENTACCELTOTAL), abnormal fetal heart ate deceleration (ABNDECELTOTAL), abnormal fetal heart rate total (ABNFHRTOTAL), and excessive uterine activity (EXUATOTAL) are the features that are most commonly recommended for improvement to avoid abnormal labor outcomes. This finding provides insights that the risk factors observed from ultrasound during labor are more important than the other risk factors.

\begin{table}[h]
\centering
\caption{Top 10 most frequently changed features in counterfactual examples for avoiding an abnormal labor outcome. The definitions of these risk factors can be found in this paper by Mamun et al. \cite{mamun2023neonatal}.}
\begin{tabular}{lc}
\hline
\textbf{Feature Name} & \textbf{Percentage of Examples (\%)} \\ \hline

ABSENTACCELTOTAL & 54.24\% \\

ABNDECELTOTAL & 49.15\% \\

ABNFHRTOTAL & 32.20\% \\

ABNVARTOTAL & 30.51\% \\

EXUATOTAL & 23.73\% \\

Fetal weight & 16.95\% \\

Labor hours & 10.17\% \\

Parity & 8.47\% \\

Bradycardia & 3.39\% \\

Short & 1.69\% \\
\hline

\end{tabular}
\label{tbl:feature_changes}
\end{table}

\begin{table*}[htbp]
\caption{Performance metrics of different classifiers on predicting abnormal delivery cases with prenatal features. All models were trained with Adam optimizer and cross-entropy loss function. All results of this table are evaluated on real and unseen test data. The unrestricted AIMEN system's performance is compared with the restricted AIMEN (R-AIMEN) systems. R-AIMEN systems set a condition on the generated data so that a minimum silhouette score is ensured among the generated data. In these experiments, the MLP\_v5 backbone was used. All models in this table were trained for up to 1000 epochs with early stopping enabled with a learning rate of 0.0001. The 8-fold cross-validation method was used in all the models in this table. F1$^+$ and F1$^-$ are the F1 scores for the abnormal class and normal class respectively. Avg F1 is the macro average F1 score of both classes.} 
\centering

\begin{tabular}{lllllllllll}
\hline
\textbf{Model}  & \textbf{Silhouette} & \textbf{Loss} &  \textbf{Accuracy} & \textbf{Sensitivity} & \textbf{Specificity} & \textbf{F1$^+$} & \textbf{F1$^-$} & \textbf{Avg F1} \\ \hline
AIMEN       & None   & 0.863         & 0.789             & 0.632                & 0.947                 & 0.750        & 0.818        & \textbf{0.784}               \\
R-AIMEN & Negative  & 0.865         & 0.763             & 0.579                & 0.947                       & 0.710        & 0.800        & 0.755               \\
R-AIMEN & Both   & 1.024         & 0.737             & 0.474                & 1.000                    & 0.643        & 0.792        & 0.717     \\ \hline         
\end{tabular}
\label{tbl:classifier_metrics}
\end{table*}

\begin{table}[htbp]
\centering
\caption{Validation loss and test loss of the AIMEN and R-AIMEN models. Three different restrictions with silhouette scores were evaluated: no restriction, restriction on the negative class, and restriction on both classes. The unrestricted AIMEN had the best distribution gap, which means the generated data was closer to the test data than other methods. Suppose, the test loss is $L_{test}$ and the average validation loss is $L_{val}$. Then, distribution gap was calculated by $\frac{L_{test} - L_{val}}{L_{val}} \times 100$.}

\begin{tabular}{llllll}
\hline
Model & Silhouette             & Best val loss  & $L_{val}$  & $L_{test}$      & Dist. gap (\%) \\ \hline
AIMEN  & None   & 0.069          & 0.134          & \textbf{0.863} & \textbf{545}      \\
R-AIMEN & Negative & 0.057          & 0.099          & 0.865          & 776               \\
R-AIMEN & Both & \textbf{0.051} & \textbf{0.077} & 1.024          & 1228   \\ \hline          
\end{tabular}

\label{tbl:dist_gap}
\end{table}

\subsection{Restricted AIMEN (R-AIMEN)}
\label{sec:raimen_results}
The default AIMEN model uses CTGAN to generate synthetic data without any restriction. On the other hand, the restricted models require the synthetic data to satisfy the condition that the average silhouette score of the two clusters (positive and negative) must increase from the previous iteration or it has to be higher than 0.3. This restriction makes the data more easily separable. However, in \tblref{classifier_metrics}, it can be seen that this restriction reduces the performance of the classifier based on the average F1 score. The reason may be that this restriction increases the distance between the distribution of the training data and the distribution of the test data because we are only using real data in the test set, and this restriction may not follow the true behavior of the data. We developed another system where a requirement of silhouette score on the synthetic data for the negative class was applied, but the positive class samples were generated freely. The goal was to increase the sensitivity of the classifier by giving the positive synthetic data more freedom than the negative synthetic data. If we look at the results, we see that the sensitivity of R-AIMEN with negative class restriction (0.579) is in fact higher than that of R-AIMEN with both class restrictions (0.474), but overall, the unrestricted AIMEN has the highest sensitivity score (0.632) among these three. The contribution of the R-AIMEN solutions is further discussed in Section~\ref{sec:raimen_discussion}.

The average F1 scores reported in \tblref{classifier_metrics} show that the unrestricted AIMEN has the highest score (0.784) among all the models. From these results, we conclude that synthetic data helps increase the performance of a model, but it is important to ensure that the distributions of the training and test data are similar after data augmentation.

\subsection{Distribution Gaps Through Validation and Test Loss}
Finally, we compared the validation and test losses to determine the distribution gap, defined as the relative difference between the average validation loss and the average test loss. In \tblref{dist_gap}, it can be seen that the distribution gap is lowest in the unrestricted AIMEN. The minimum best validation loss or average validation loss is achieved when the silhouette score is applied to both classes. However, better validation loss does not necessarily translate to better test loss. Applying a silhouette score restriction makes the synthetic data more easily separable, hence the validation loss is lower. However, in this way, the model fails to learn some of the distinctive features of the data, as the restricted synthetic data does not follow the true distribution of the data because the real data does not have to be easily separable in general. That is why, despite better validation metrics, R-AIMEN models could not achieve test metrics as good as AIMEN's. Hence, the distribution gap is lowest in the unrestricted AIMEN.

\begin{figure}[tbh!]
    \centering
        \includegraphics[width=0.98\textwidth]{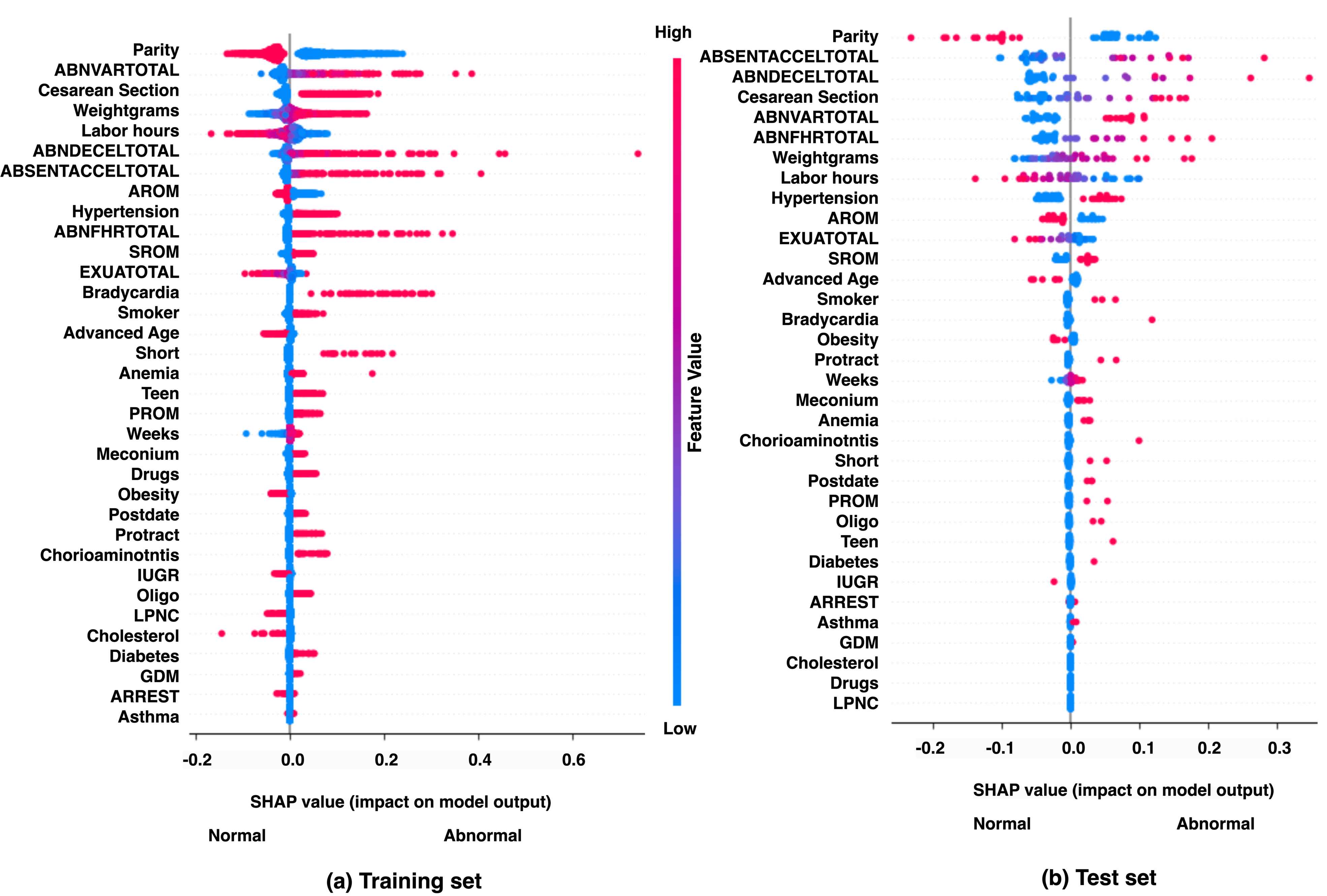}
    \caption{SHAP values for 34 input features on 1385 real training data and 38 real test data using the AIMEN system (MLP\_v5 backbone). The SHAP plots suggest that abnormal heart rate (risk factors ABNVARTOTAL, ABNDECELTOTAL, ABSENTACCELTOTAL, ABNFHRTOTAL) or breathing conditions (Bradycardia) can increase the risk of abnormal delivery, which is consistent with the clinical understanding \cite{nihWhatRisk}. The SHAP plot further suggests that a mother with low parity, such as nulliparous mothers may have a comparatively higher risk than multiparous mothers. Although the effect of parity on the labor outcome for young mothers is unestablished in previous studies, nulliparous mothers with advanced age face increased risk of complicated pregnancies according to prior research \cite{shechter2020does}, supporting our findings.}
    \label{fig:shap_figures}
\end{figure}

\subsection{SHAP Values for Training and Test Datasets}
To understand more about the AIMEN system's method of decision-making, we have plotted the SHAP (SHapley Additive exPlanations) values \cite{lundberg2017unified} on 1385 real training cases and 38 real test cases in \figref{shap_figures}. On this test set, the macro average F1 score of the classifier was 0.78, and the AUROC was 0.77. From the SHAP values of the training and test sets, we observed that cases with low parity, high abnormal acceleration, high abnormal deceleration, cesarean section, and high abnormal variability are some of the combinations that influenced our system to predict a case as abnormal. However, excessive uterine contractions or advanced maternal age did not usually influence our system to make abnormal class predictions. It needs to be noted that the SHAP values are calculated based on the model's predictions instead of the ground-truth labels. So, these findings do not necessarily mean that the relationships of these features with the outcome will be the same for ground-truth observations. Nonetheless, it can give us an intuition about how our prediction model works and provide us with directions on how to improve the system in the future. As illustrated in \figref{shap_figures}, the SHAP analysis highlights that abnormal fetal heart rate patterns—captured through risk indicators such as ABNVARTOTAL, ABNDECELTOTAL, ABSENTACCELTOTAL, and ABNFHRTOTAL—along with irregular breathing conditions such as bradycardia, are strongly associated with an elevated risk of abnormal delivery outcomes. These findings are consistent with well-established clinical evidence~\cite{nihWhatRisk}, which recognizes fetal heart rate irregularities as key indicators of fetal distress and delivery complications. Furthermore, the SHAP plots reveal that low maternal parity, particularly in nulliparous women, is associated with a comparatively higher likelihood of abnormal delivery when contrasted with multiparous mothers. While the relationship between parity and labor outcomes in younger mothers remains inconclusive, prior research has demonstrated that nulliparous women of advanced maternal age exhibit an increased risk of pregnancy-related complications~\cite{shechter2020does}. These observations reinforce that the AIMEN system successfully captures clinically meaningful RFs consistent with established medical understanding.

\subsection{Additional Experiments - Prediction of Individual Outcomes}

The results in \tblref{separate_outcomes} demonstrate that the AIMEN system exhibits varying performance across different individual outcome predictions. Among the three outcomes, Apgar score at 1 minute $\leq$ 3, umbilical cord pH $\leq$ 7.05, and cerebral palsy (CP), the model achieved the strongest predictive performance for CP, with an accuracy of 0.778, sensitivity of 0.611, average F1 score of 0.763, and AUROC of 0.880. These results indicate that the AIMEN framework is more effective in identifying cases associated with long-term neurological complications than short-term post-delivery measures such as Apgar score or pH level. In contrast, the relatively low sensitivities for Apgar (0.286) and pH (0.233) suggest that these outcomes are harder to predict from the available features, despite their high specificities and positive predictive values. This imbalance implies that while the model confidently identifies normal cases, it struggles to detect true abnormal instances for these short-term indicators. Future work will focus on developing specialized models tailored to individual outcomes such as pH, Apgar scores at 1 minute and 5th minute, NICU admission, neonatal encephalopathy, and neonatal hemorrhage, to enhance predictive sensitivity and improve clinical utility across diverse neonatal endpoints.

\begin{table}[h]
\centering
\caption{Additional experiments for prediction of individual outcomes: CP, Apgar score at 1 minute, and umbilical cord pH. The AIMEN system with CTGAN augmentation was used for balancing and augmenting the training data. A balanced test set containing only real data points was separated before any balancing and augmentation to prevent data leakage. The experiment was repeated three times, and average performance metrics are reported below. The MLP\_v5 backbone was used for the experiments in this table. Results suggest that when predicting individual outcomes, AIMEN performs well for predicting CP, whereas the predictions of Apgar at 1 minute and pH produce low sensitivity scores. Boldfaced values represent the highest performance metric among the three different outcomes.}
\begin{tabular}{rccccccccc}
\hline
\textbf{Outcome}                  & \textbf{Acc}   & \textbf{Sens}  & \textbf{Spec}  & \textbf{PPV}   & \textbf{NPV}   & \textbf{F1$^+$}   & \textbf{F1$^-$}   & \textbf{Avg F1} & \textbf{AUROC} \\ \hline
Apgar at 1 minute $\leq$ 3 & 0.631          & 0.286          & 0.976          & 0.958          & 0.584          & 0.405          & 0.728          & 0.566           & 0.712          \\
pH $\leq$ 7.05              & 0.617          & 0.233          & \textbf{1.000} & \textbf{1.000} & 0.569          & 0.362          & 0.724          & 0.543           & 0.710          \\
CP                                & \textbf{0.778} & \textbf{0.611} & 0.944          & 0.933          & \textbf{0.724} & \textbf{0.712} & \textbf{0.814} & \textbf{0.763}  & \textbf{0.880} \\ \hline
\end{tabular}
\label{tbl:separate_outcomes}
\end{table}

\section{Discussion}
\subsection{Key Contributions}
This work introduced AIMEN, an AI pipeline intended to support clinical decision-making during delivery with several noteworthy properties. AIMEN addresses data challenges in neonatal health monitoring with a synthetic-data-augmented ensemble framework and provides explainable AI through CE. Our experiments demonstrated that CTGAN-based augmentation provides more effective support for downstream classification than ADASYN. Among the tested configurations, the unrestricted AIMEN with an MLP\_v5 backbone achieved the best overall performance (macro average F1 = 0.784, sensitivity = 0.632, specificity = 0.947). Restricted AIMEN variants that enforced cluster separability reduced validation loss but performed worse on the test set, highlighting that preserving distributional consistency between training and test data is more important than artificially improving separability. CEs further showed that AIMEN can generate meaningful explanations with low distance (0.33) and sparsity (2.5 feature changes), demonstrating its potential for both predictive accuracy and clinical interpretability.

AIMEN outperforms the baselines such as XGBoost, TabNet, \danets, LightGBM, and MLPs. Through SHAP analysis and counterfactual analysis, we have also provided insights into which features may contribute most toward abnormal labor outcomes. These insights pave the way for future research and development for clinically deployable solutions. Finally, through additional experiments, we also showed that AIMEN is able to predict CP with an F1 score of 0.763 and an accuracy of 0.778.

\subsection{Insights into the Explainability}
Explainability analysis using SHAP revealed that AIMEN relies on clinically meaningful features when predicting abnormal outcomes. SHAP values consistently identified parity, abnormal accelerations and decelerations, cesarean section, and abnormal variability as influential predictors. Other factors, such as uterine contractions or maternal age, had less impact on the model’s decisions, indicating that AIMEN prioritizes a clinically plausible but selective subset of features. CEs complemented these global insights by showing that relatively few changes (average sparsity of 2.5) were sufficient to flip predictions, providing actionable recommendations rather than opaque risk scores. 

The SHAP summary plots also confirmed consistency between training and test sets, with absent accelerations and abnormal decelerations ranking highest in importance. These features correspond to established clinical risk factors such as fetal hypoxia or cord compression, which align with obstetric guidelines for intervention. Additional features such as low birth weight, prolonged labor, and hypertension contributed to the model’s decisions, further supporting its clinical relevance. Together, SHAP and counterfactual analyses provide complementary perspectives: SHAP values explain global model behavior, while counterfactuals offer local, actionable insights for individual cases.

Moreover, the sparsity values of our CEs are in the optimal range according to previous studies~\cite{ijcai2021p609keane2021if}, as the users of explainable AI systems prefer counterfactuals that require changes in 2-3 features instead of 1-feature or higher than 3 features \cite{forster2020fostering, ijcai2021p609keane2021if, forster2021capturing, dodge2019explaining, lim2009and}.

\subsection{Contribution of the R-AIMEN Models}
\label{sec:raimen_discussion}
The R-AIMEN variant extends AIMEN by incorporating a silhouette-score-based filtering mechanism that retains synthetic samples closely aligned with the majority of datapoints within each class. This step excludes data points located far from the main class clusters, reducing the influence of low-density or ambiguous regions in the feature space. As the silhouette score is making data more separable by forcing the generated datapoints to be close to the clusters, it will improve the performance on the training and the validation set, as shown in \tblref{dist_gap}.
However, adding silhouette score filtering may make the data less realistic as it removes certain groups of data points for being too far from the clusters and thus may not represent the test set distribution properly. Hence, in our experiments, AIMEN outperformed R-AIMEN methods on the test set as shown in \tblref{dist_gap}.

The design of filtering generated datapoints based on silhouette scores was inspired by the study by the approach by Zhai et al. \cite{zhai2022binary} that proposed using silhouette scores to assess the quality of synthetic samples.
That study, however, did not include an ablation analysis to evaluate the effect of disabling the filtering step. In our work, we addressed this limitation by comparing both AIMEN (without filtering) and R-AIMEN (with silhouette-based filtering) to examine the trade-offs empirically. We found that R-AIMEN improved performance on training and validation data, suggesting better alignment between real and synthetic distributions, but did not yield higher accuracy on the external test set. From a clinical perspective, this filtering approach appears most suitable when stable, class-consistent learning is prioritized over marginal gains in generalization. Overall, the method adds an extra layer of reliability for understanding model behavior and ensures that synthetic augmentation remains centered around clinically meaningful data clusters.

\subsection{Statistical Significance Tests and Confidence Interval Analysis}
\label{sec:statistical_analysis}
To rigorously evaluate whether performance differences among the five models are statistically meaningful, we employed the Friedman test~\cite{friedman1937use} followed by the Nemenyi post-hoc test~\cite{nemenyi1963distribution} for pairwise comparisons. This non-parametric approach was chosen over parametric alternatives (e.g., repeated-measures ANOVA) as the Friedman and Nemenyi tests do not require normality assumptions, hence they are the most compatible tests for our experimental setting.

The Friedman test assesses whether the distribution of ranks assigned to models across trials differs significantly from what would be expected under the null hypothesis of equal performance. Effect size was quantified using Kendall's $W$ (coefficient of concordance), where $W = 0$ indicates no agreement in rankings across trials and $W = 1$ indicates perfect agreement. Upon finding a significant Friedman result, pairwise comparisons were conducted using the Nemenyi post-hoc test, which controls the family-wise error rate across all model pairs.

The Friedman test revealed statistically significant differences among the five models on both accuracy ($\chi^2 = 13.96$, $p = 0.0074$) and F1 score ($\chi^2 = 14.56$, $p = 0.0057$), as seen in \tblref{statistical_analysis}. The Kendall's $W$ values of 0.698 and 0.728 for accuracy and F1, respectively, indicate strong concordance across trials, meaning that the relative ranking of models was highly consistent and not an artifact of any single trial.

Nemenyi post-hoc comparisons revealed that the primary driver of the overall Friedman significance is the infereior performance of TabNet, as seen in \tblref{statistical_analysis}. For accuracy, TabNet was significantly outperformed by XGBoost and AIMEN ($p < 0.05$). For the F1 score, TabNet was significantly outperformed by both XGBoost and LightGBM ($p < 0.05$), with a marginal but non-significant trend against AIMEN ($p = 0.070$). Among the remaining four models, no statistically significant pairwise differences were detected. AIMEN achieved the highest mean accuracy ($0.753 \pm 0.044$) and F1 score ($0.737 \pm 0.050$) across all models, with all pairwise comparisons against XGBoost, LightGBM, and \danets returning $p > 0.05$, as seen in \tblref{statistical_analysis}.

\begin{table}[h]
\centering
\caption{Friedman test results and Nemenyi post-hoc pairwise p-values for accuracy and macro average F1 score across 5 trials. $W$ denotes Kendall's coefficient of concordance. Pairwise p-values below 0.05 are in bold.}
\label{tbl:statistical_analysis}
\begin{tabular}{p{0.5in}lccccc}
\hline
\textbf{Metric} & & \danets & XGBoost & LightGBM & TabNet & AIMEN \\
\hline
\multirow{7}{0.5in}{\textbf{Accuracy p-values}}
 & $\chi^2 = 13.9583,\ p = 0.0074,\ W = 0.6979$ & & & & & \\[4pt]
 & \danets    & ---    & 0.4339 & 0.6912 & 0.6277 & 0.6277 \\
 & XGBoost  & 0.4339 & ---    & 0.9946 & \textbf{0.0166} & 0.9982 \\
 & LightGBM & 0.6912 & 0.9946 & ---    & 0.0539 & 1.0000 \\
 & TabNet   & 0.6277 & \textbf{0.0166} & 0.0539 & --- & \textbf{0.0409} \\
 & AIMEN    & 0.6277 & 0.9982 & 1.0000 & \textbf{0.0409} & --- \\
\hline
\multirow{7}{0.5in}{\textbf{F1 Score p-values}}
 & $\chi^2 = 14.5567,\ p = 0.0057,\ W = 0.7278$ & & & & & \\[4pt]
 & \danets    & ---    & 0.3735 & 0.4339 & 0.6912 & 0.6912 \\
 & XGBoost  & 0.3735 & ---    & 1.0000 & \textbf{0.0166} & 0.9874 \\
 & LightGBM & 0.4339 & 1.0000 & ---    & \textbf{0.0227} & 0.9946 \\
 & TabNet   & 0.6912 & \textbf{0.0166} & \textbf{0.0227} & --- & 0.0703 \\
 & AIMEN    & 0.6912 & 0.9874 & 0.9946 & 0.0703 & --- \\
\hline
\end{tabular}
\end{table}

Taken together, these results support the conclusion that AIMEN performs competitively with the strongest baseline methods (XGBoost and LightGBM), achieving the highest numerical performance on both metrics while remaining statistically indistinguishable from them. Future work involving a larger number of trials would provide greater statistical power to formally distinguish AIMEN from XGBoost and LightGBM, where a meaningful performance trend is already observable.

To further assess the robustness of each model, 95\% confidence intervals (CIs) were computed over five independent runs (\tblref{baseline_vs_aimen_combined}). AIMEN achieved a narrow CI range for both accuracy ([0.698, 0.807]) and macro average F1 score ([0.675, 0.800]), reflecting consistent performance across trials. In contrast, the multilayer perceptrons showed broader intervals, particularly MLP-1 and MLP-4, indicating higher variability and less generalization stability.

XGBoost and LightGBM demonstrated relatively tight confidence intervals but remained below AIMEN’s mean accuracy and macro average F1 score. TabNet’s wide CI overlap with the lower spectrum of all models underscores its unstable learning behavior on this dataset.

The narrow bounds of AIMEN’s confidence intervals confirm that its superior average scores are statistically reliable rather than random fluctuations, supporting the robustness of the proposed framework.

\section{Limitations and Future Works}
It is very important to identify labor risks as early as possible to prevent or mitigate adverse labor outcomes. Our study makes novel and significant contributions toward this goal. We propose a method to train neural networks for classification problems with small datasets. However, it is difficult to properly evaluate the effect of an RF on an outcome without an RCT or an observational study with a large dataset. One challenge is that RCTs may not always be feasible or ethical in the setting of intrapartum care. Our study proposes to address this issue by providing CE for abnormal outcomes, which gives an idea of what factors would have to be different for a normal outcome. This study uses only one counterfactual generation method. The scope of the study for classifiers is limited to fully connected neural networks and how to improve their capacity with ADASYN and CTGAN-based data augmentations.

The small dataset remains a key limitation, as it may restrict the model’s ability to capture the full variability of labor characteristics across diverse populations. Although data augmentation improved performance on the real test data, synthetic samples cannot entirely replace the diversity and nuances of real-world cases. Another important consideration is the translation of this approach into prospective clinical use, which would require further validation with larger, multi-institutional datasets, integration with electronic health records, and evaluation of real-time feasibility. Nonetheless, the study establishes a solid foundation for future work by showing that meaningful predictive insights and interpretable CEs can be derived even from small datasets, paving the way toward early and explainable labor risk identification. 

In the future, we plan to perform a large-scale user study for better evaluation and improvement of our system. Additionally, we aim to fine-tune the system for other adverse outcomes, such as NICU admission and characteristics of the neonate shortly after birth. Moreover, integrating an option to choose from multiple counterfactual generation methods may better assist physicians by providing solutions from various sources.
\section{Conclusions}
Classification with tabular data is challenging, especially when the output classes are highly imbalanced. Our study explored different methods to predict the high risk of adverse labor outcomes and provide CE. We connected neonatal risk modeling, tabular data classification, and CE to address this important problem. Our work overcomes the challenges of limited and imbalanced data by employing generative models for data balancing and augmentation. It highlights the drawbacks of imposing restrictions on the generated data based on separability. Our experiments demonstrate that a systematically chosen neural network supported by an unrestricted CTGAN can outperform the models not supported by a CTGAN and those supported by a restricted CTGAN. Our method predicts the high risk of adverse labor outcomes with a positive class F1 score of 0.75 and an average F1 score of 0.784. Neonatal health risk prediction with ML/AI is an understudied topic, and we invite researchers to contribute to this important field.

\section*{Acknowledgment}
This work was supported in part by WearTech Center, an applied research center owned and operated by the Partnership for Economic Innovation. Any opinions, findings, conclusions, or recommendations expressed in this material are those of the authors and do not necessarily reflect the views of the funding organization.

\bibliographystyle{ACM-Reference-Format}
\bibliography{references}

\end{document}